\documentclass[conference]{IEEEtran}

\usepackage{fancyhdr}  % Package for custom headers/footers

\usepackage{xcolor}
\usepackage{soul}
\sethlcolor{yellow} % Set highlight color to yellow

\usepackage{multirow}
\usepackage{makecell}
\usepackage{amsmath}
\usepackage{hyperref}

\usepackage{graphicx} 

\usepackage{cite}

\ifCLASSINFOpdf

\else

\fi

\usepackage{array}

\ifCLASSOPTIONcompsoc
 \usepackage[caption=false,font=normalsize,labelfont=sf,textfont=sf]{subfig}
\else
 \usepackage[caption=false,font=footnotesize]{subfig}
\fi

\begin{document}

\thispagestyle{empty}  % Suppress page number on the first page

\title{How Different Tokenization Algorithms Impact LLMs and Transformer Models for Binary Code Analysis }

\newpage  % Start numbering from the second page

\author{\IEEEauthorblockN{\IEEEauthorrefmark{1}Ahmed Mostafa,
\IEEEauthorrefmark{1}Raisul Arefin,
Samuel Mulder}
\IEEEauthorblockA{\IEEEauthorrefmark{1}These authors contributed equally}

\IEEEauthorblockA{Auburn University \\
\{aim0008, ran0013, szm0211\}@auburn.edu}

% \IEEEauthorblockA{Department of Computer Science and Software Engineering\\
% Auburn University}
}

\IEEEoverridecommandlockouts
\makeatletter\def\@IEEEpubidpullup{6.5\baselineskip}\makeatother 
\IEEEpubid{\parbox{\columnwidth}{
		Workshop on Binary Analysis Research (BAR) 2025 \\
		28 February 2025, San Diego, CA, USA \\
		ISBN 979-8-9919276-4-2 \\ 
		https://dx.doi.org/10.14722/bar.2025.23013 \\   
		www.ndss-symposium.org
}
\hspace{\columnsep}\makebox[\columnwidth]{}}

% make the title area
\maketitle

% \end{abstract}
\begin{abstract}
Tokenization is fundamental in assembly code analysis, impacting intrinsic characteristics like vocabulary size, semantic coverage, and extrinsic performance in downstream tasks. Despite its significance, tokenization in the context of assembly code remains an underexplored area. This study aims to address this gap by evaluating the intrinsic properties of Natural Language Processing (NLP) tokenization models and parameter choices, such as vocabulary size. We explore preprocessing customization options and pre-tokenization rules tailored to the unique characteristics of assembly code. Additionally, we assess their impact on downstream tasks like function signature prediction---a critical problem in binary code analysis.

To this end, we conduct a thorough study on various tokenization models, systematically analyzing their efficiency in encoding assembly instructions and capturing semantic nuances. Through intrinsic evaluations, we compare tokenizers based on tokenization efficiency, vocabulary compression, and representational fidelity for assembly code. Using state-of-the-art pre-trained models such as the decoder-only Large Language Model (LLM) Llama 3.2, the encoder-only transformer BERT, and the encoder-decoder model BART, we evaluate the effectiveness of these tokenizers across multiple performance metrics. Preliminary findings indicate that tokenizer choice significantly influences downstream performance, with intrinsic metrics providing partial but incomplete predictability of extrinsic evaluation outcomes. These results reveal complex trade-offs between intrinsic tokenizer properties and their utility in practical assembly code tasks. Ultimately, this study provides valuable insights into optimizing tokenization models for low-level code analysis, contributing to the robustness and scalability of Natural Language Model (NLM)-based binary analysis workflows.
\end{abstract}

\IEEEpeerreviewmaketitle

\section{Introduction}
Tokenization is critical in transforming raw input data into structured representations, a process of utmost importance for Machine Learning (ML) and NLM model tasks~\cite{sennrich-etal-2016-neural, gu2022uniasm, gao2021lightweight}. While tokenization strategies have been studied extensively for natural~\cite{ali-etal-2024-tokenizer} and high-level programming languages~\cite{dagan2024getting}, assembly code presents unique challenges due to its low-level operations, diverse instruction sets, and non-standardized syntax across architectures. These challenges highlight the need for specialized tokenization techniques that effectively capture assembly code's structural and semantic intricacies~\cite{gu2022uniasm}. Despite its importance, the role of tokenization in assembly code processing remains underexplored, particularly in its impact on downstream tasks involving modern NLMs.

Recent research underscores the significant influence of tokenization on NLM model performance. Studies like Ali et al.\cite{ali-etal-2024-tokenizer} demonstrate that tokenization methods affect the model's efficiency and ability to generalize across NLP tasks. Additionally, Dagan et al.\cite{dagan2024getting} emphasize the critical role of tokenizers in domain adaptation and fine-tuning, showing that pre-tokenization schemes and vocabulary size significantly impact model compression rates, training efficiency, and downstream performance. For binary code analysis, tokenization has also played a pivotal role in downstream tasks like binary similarity detection UniASM~\cite{gu2022uniasm} and function name reassignment Gao et al.~\cite{gao2021lightweight}. However, existing tokenization methods often fall short when applied to assembly code, primarily due to their reliance on token patterns optimized for natural language or high-level code, leading to suboptimal results in binary-focused applications.

Assembly code's reliance on hardware-specific instructions and lack of high-level abstractions complicates the creation of structured representations. CP-BCS~\cite{ye-etal-2023-cp} addresses this challenge by integrating Control Flow Graphs (CFGs) and pseudo code to bridge the semantic gap between assembly and human-readable summaries. Particularly in stripped binaries, advanced tokenization approaches, such as bidirectional instruction-level CFGs and pseudo-code refinement, prove essential for capturing execution behavior and logic semantics. CP-BCS~\cite{ye-etal-2023-cp} work emphasizes the need for tailored preprocessing Gao et al.~\cite{gao2021lightweight} and CP-BCS~\cite{ye-etal-2023-cp}, tokenization, and post-tokenization methods in assembly code analysis.

Tokenizers are crucial tools in processing assembly code for NLP tasks in binary analysis, as poorly designed tokenization strategies can significantly hinder model performance. Despite their importance, no comprehensive study has systematically evaluated intrinsic and extrinsic tokenizer performance in assembly code. This study addresses these limitations by systematically evaluating tokenization strategies for assembly code. We focus on the performance of various tokenizers and tokenization methods when applied to decoder-only models, such as Llama 3.2 1B parameters (the model can be downloaded from \href{https://huggingface.co/meta-llama/Llama-3.2-1B-Instruct}{here}), encoder-only models, such as BERT~\cite{devlin-etal-2019-bert}, and encoder-decoder models such as BART~\cite{lewis2019bart}. Specifically, our contributions are as follows:

\begin{itemize}
    \item We analyze intrinsic tokenizer performance in detail, assessing its ability to encode assembly instructions and components effectively.
    \item We evaluate extrinsic tokenizer performance by examining its impact on downstream tasks.
      \item We evaluate the impact of preprocessing the instructions before tokenization for downstream tasks.
\end{itemize}

This research contributes to the field by filling a critical gap in understanding tokenization for assembly code. It offers a framework for evaluating and optimizing tokenization strategies tailored to binary program analysis. By leveraging state-of-the-art models and domain-specific datasets, we provide actionable insights for developing more effective tokenization methods, ultimately advancing the capabilities of NLMs in binary analysis.

\section{Background}
Tokenization is crucial in natural language processing and binary analysis, which bridges raw data and machine understanding. It involves segmenting text or binary code into smaller units, known as tokens, enabling efficient processing by machine learning and large language models. The choice of tokenization strategy significantly influences model performance~\cite{ali-etal-2024-tokenizer} mainly when dealing with specialized languages like assembly code.

\subsection{Tokenization Algorithms}

One weakness of using just words as tokens is that the vocabulary size will be unmanageable. If we have a maximum size limit of the vocabulary, there will be a lot of out-of-vocabulary (OOV) words. One solution might be to tokenize text based on characters. However, that would not generate meaningful tokens, and the number of tokens generated would be very large, even from a shorter text. 
The subword-based tokenization algorithms try to maintain a balance between these two approaches. This approach generally tries to keep frequent smaller words without splitting in the vocabulary. For example, the word ``eat" might not be split, but ``eating" might be split into ``eat" and ``ing". The subword-based tokenization methods we will be using are:

\subsubsection{\textbf{WordPiece}}
WordPiece algorithm~\cite{schuster2012japanese} is a subword-based tokenization formula. It was developed by Google to pretrain BERT and has been reused by other popular models like DistilBERT, MobileBERT, Funnel Transformers, and MPNET.

The WordPiece vocabulary is initialized with the special tokens and the initial alphabet. The initial alphabet is produced from the corpus. It first splits all the words in the corpus into subwords. For example, the word ``one" is broken into: `o’, `\#\#n', and `\#\#e'. Here, `o' is different from the other two alphabets because it is at the beginning of a word. In short, the initial vocabulary contains all the initial letters of a word and all the other letters preceded by the prefix `\#\#’. Then, these subwords are merged based on a rule to create longer subwords or whole words. This process is repeated until the vocabulary size is complete.
The merging rule:
For all the token pairs in the current vocabulary, a score is calculated according to the following formula: 

{\scriptsize
\[
score= \frac{ \text{Frequency of pair} }{ \text{Frequency of  first element} \times \text{Frequency of second element} }
\]
}

The pair of tokens with the highest score is merged and added to the vocabulary. Then, the process is repeated until the desired vocabulary size is reached.

During tokenizing new words, WordPiece looks for the longest match in the vocabulary. If the whole word is absent in the vocabulary, the longest match is used to split the word. For example, while tokenizing ``cats", if "cats" is not present in the vocabulary but ``cat" is, it will tokenize as ``cat" and `\#\#s'. 

\subsubsection{\textbf{Byte Pair Encoding}}
BPE~\cite{sennrich2015neural} is a data compression technique adapted as a subword tokenization method for natural language processing tasks. It was originally designed for compressing text and later used by models like GPT, GPT-2, BART, and DeBERTA.

The token selection method for vocabulary building is very similar to WordPiece. The major difference is how the score of each pair is calculated before merging. BPE starts with splitting the corpus into characters. So, the vocabulary will start with all the  ASCII characters, at the very least, and probably some Unicode characters. Then, it will find the most frequent pair instead of using the equation like in WordPiece. The most frequent pair is merged and added to the vocabulary. This method is repeated until the desired vocabulary size is reached.

For tokenizing new words, BPE uses both the vocabulary and merging rules that it learned during vocabulary production. For example, for tokenizing "cats", it will first split the word into characters like 'c’, 'a’, 't’, and 's’. Then, it will use the learned merged rules to merge them and form the longest token possible.

\subsubsection{\textbf{Unigram}}
The Unigram model~\cite{kudo-2018-subword} is a language model that takes a different approach to building its vocabulary than algorithms like WordPiece and BPE. It starts with a large vocabulary and gradually trims it down. Unigram prunes tokens based on how much they impact the model’s likelihood over the entire corpus. In each iteration, Unigram calculates a loss. This loss is computed by tokenizing every word in the corpus, using the current vocabulary and the Unigram model determined by the frequencies of each token in the corpus. Subsequently, it evaluates the potential increase in this overall loss for each symbol in the vocabulary if that symbol were to be eliminated. Then, it removes the percentage of tokens whose log increase is the lowest. This process is iterated until the desired vocabulary size is obtained.

Tokenizing a new word involves examining every possible segmentation of the word into tokens. Each segmentation is evaluated by calculating the probability of that specific sequence according to the Unigram model. The general idea is to split a word into the least number of tokens possible.

Unigram is used in SentencePiece, which is the tokenization algorithm used by popular models like AlBERT, T5, mBART, Big Bird, and XLNet.

\subsection{Related Works}

\subsubsection{Preprocessing}

Preprocessing text before tokenization involves modifying the original input in such a way that better fits the target task. For example,  converting all text to lowercase to ensure consistency, removing punctuation and special characters, etc. In the case of binary analysis, preprocessing is done a little differently than natural language. 

Some tools ignore all the numeric values, such as Escalada et al. ~\cite{escalada2021improving}, TypeMiner~\cite{maier2019typeminer}. Cati~\cite{chen2020cati} and Stride ~\cite{green2024stride} keep relatively small numeric values but remove large numbers.

Palmtree ~\cite{li2021palmtree}  performed a study for instruction representation learning. The authors preprocessed the code by replacing strings and large numeric values with special tokens. However, they kept the smaller numeric values as they often convey important information about accessed local variables, function arguments, or data structure fields. 

Gao et al.~\cite{gao2021lightweight} and CP-BCS~\cite{ye-etal-2023-cp} performed an instruction normalization method to mitigate data sparsity and OOV issues in binary analysis by simplifying instruction representation. Key steps include retaining mnemonics and registers, generalizing constants, and substituting function addresses and local jumps with placeholder tokens. This approach improves model generalization and learning efficiency.

\subsubsection{Tokenization}

Tokenization is the process of splitting text into smaller units called tokens, which can be words, subwords, or even characters. This is a crucial step in preparing text for NLP models, as it transforms raw text into a structured format that the models can understand and analyze. There are various ways binary analysis tools perform tokenization:

\textbf{Learning-based Encoding:} SnowWhite ~\cite{lehmann2022finding} is a machine learning approach to predict type information from stripped binaries. They analyzed their dataset and found that the number of unique tokens was vast due to the prevalence of numeric values. To keep the vocabulary of their tokenizer feasible, they developed a subword model tokenizer based on BPE.

Karampatsis et al. \cite{karampatsis2020big} performed an empirical study to find the best way to tokenize source code. Source codes are rich with identifiers, which can cause the vocabulary to explode. They came up with the idea of using character subsequences of tokens (subword units) to reduce the final vocabulary size. 

\textbf{Raw Bytes Encoding:} Some tools encode the instructions as raw byte encoding and feed that to the NLP model. The one-hot encoding algorithm commonly uses this scheme. A byte consists of 8 bits and offers a range of 256. A vector of length 256 with one active dimension effectively encodes bytes as vectors. A few tools that pass similar input to NLP models are StateFormer~\cite{pei2021stateformer}, DEEPVSA~\cite{guo2019deepvsa}, EKLAVYA~\cite{chua2017neural} and~\cite{shin2015recognizing}.

\textbf{Instruction-level tokenization:} UniASM~\cite{gu2022uniasm} evaluated the BPE, WordPiece, and three instruction-level (Full, Half, and Piece Instruction) tokenization algorithms to assess their effectiveness in binary code similarity detection tasks. The evaluation results demonstrated that the Full-Instruction tokenization method, which treats a single instruction as a token, consistently outperformed the other approaches by preserving instructions' structural and semantic integrity. Despite the effectiveness of the Full-Instruction tokenization technique, it can result in a more extensive vocabulary and is more susceptible to OOV issues.

StateFormer \cite{pei2021stateformer} took a different approach to handling numeric values. They used Neural Arithmetic Unit (NAU) ~\cite{madsen2020neural}, embeddings produced by which are supposed to capture the semantics of numerical values involved in arithmetic operations.

\section{Approach}

\subsection{Tokenizers}

To evaluate the impact of tokenizers on model performance, we conducted an ablation study focusing on the pre-trained models: the decoder-only Llama 3.2 1B parameters, the encoder-only BERT, and the encoder-decoder BART-Base~\cite{lewis2019bart} model. Specifically, we created a diverse dataset for training customized tokenizers and models, including 80,000 disassembled C functions for model training and 20,000 disassembled functions for testing. We trained the models for each tokenizer while fixing the remaining configurations, such as datasets, training procedures, and hyperparameters. This controlled setup enabled us to isolate and quantify each tokenizer's effect on the models' downstream performance.

\begin{figure*}[!h]
    \centering
    \includegraphics[width=0.85\textwidth]{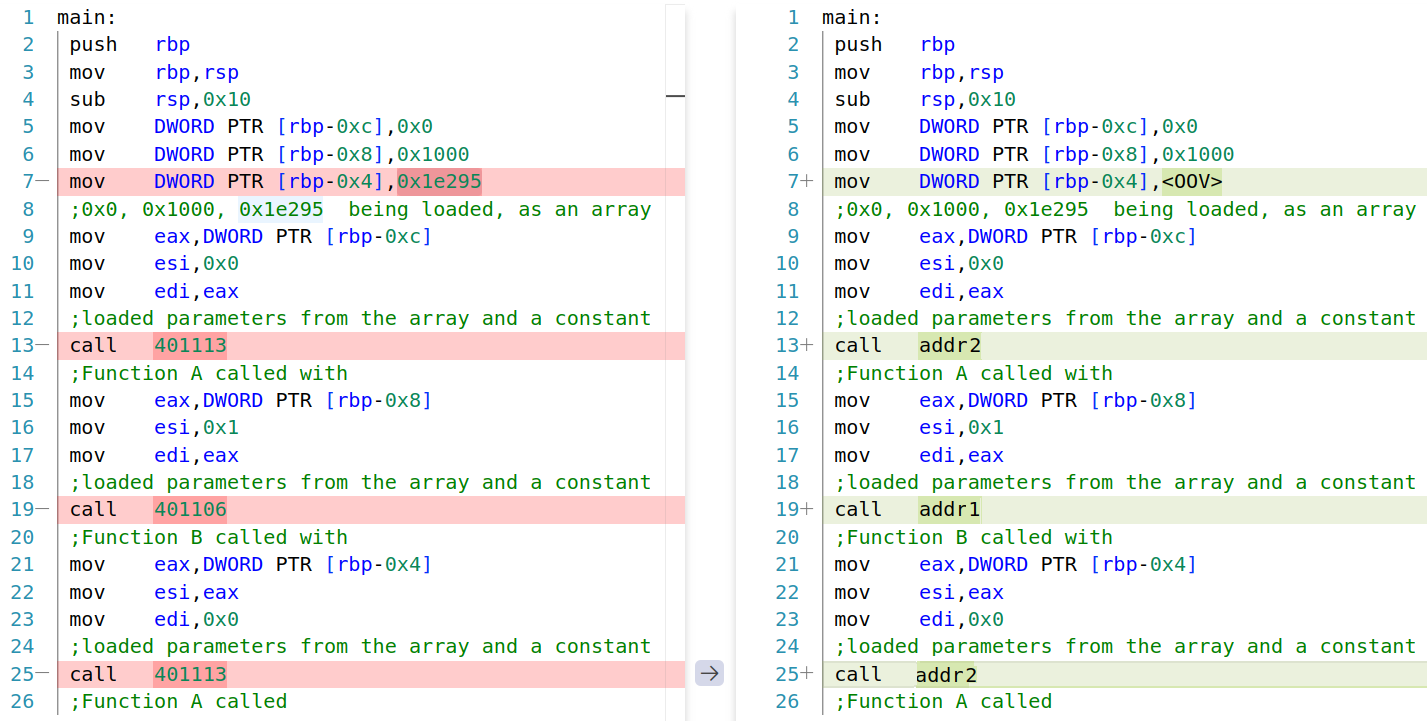} 
    \caption{An example of address-to-sequential-identifiers preprocessing. The code on the left represents the original code before preprocessing, while the code on the right shows the result after preprocessing. }
    \label{fig:address_to_words}
\end{figure*}

Our study uses the \textcolor{blue}{\href{https://huggingface.co/learn/nlp-course/en/chapter6/8}{Hugging Face tokenizer library}} to implement three well-established tokenization algorithms: BPE, Unigram, and WordPiece. Each tokenization algorithm was tested with three vocabulary sizes: 3K, 25K, and 35K. Additionally, only the Llama 3.2 model was tested with a 128K vocabulary size across all tokenization algorithms on the function signature prediction downstream task. The 128K vocabulary size is comparable to the Llama 3.2 model's default tokenizer's vocabulary size.

To further evaluate the impact of the code preprocessing, each of these tokenizers was trained on two versions of the dataset: the default disassembly without any preprocessing and the other customized using a preprocessing method. The preprocessing method is discussed in detail in the subsection (\textbf{\textit{C. Dataset}}) from this section.

In addition, the base tokenizer that comes pre-trained with the Llama 3.2, BERT, and BART models was evaluated on both versions of the dataset without any customization or additional training. This, combined with assessing the trained tokenizers, resulted in \textbf{(86 models)} across the (Llama 3.2, BERT, and BART) models, with three tokenization algorithms (BPE, Unigram, and WordPiece) and four vocabulary sizes (3K, 25K, 35K, and 128K). This experimental setup ensures a comprehensive and comparative evaluation of the Llama 3.2, BERT, and BART models. This design allows us to systematically analyze the effects of tokenization algorithms, vocabulary sizes, and dataset preprocessing on downstream model performance.

The choice of vocabulary sizes was driven by the need to balance efficiency and expressiveness. Smaller vocabularies, such as 3K, are expected to reduce memory and computational overhead while increasing sequence length due to more granular subword segmentation. On the other hand, larger vocabularies, like 35K and 128K, may better capture semantic and syntactic patterns by encoding longer subwords, reducing sequence length but increasing memory usage. The 25K vocabulary serves as a middle ground to assess trade-offs between granularity and model efficiency. Additionally, the 128K vocabulary size was included to evaluate the tokenizers' ability to handle an extensive vocabulary, capturing a wide range of tokens and potential rare subwords, which could be beneficial for highly complex and diverse datasets.

By varying the tokenization algorithm and vocabulary sizes, we aim to analyze the effect of tokenization granularity on downstream model performance, particularly on the decoder-only, encoder-only, and encoder-decoder models. This approach enables us to identify the optimal tokenizer configuration for assembly code analysis tailored to a specific downstream task while accounting for algorithmic and vocabulary design choices. The tokenizers' configurations are described in \hyperref[Tokenizer Hyper-Parameters]{Appendix A}.

\subsection{Preprocessing}

NLP models often struggle with understanding and processing numbers effectively because they are primarily trained on textual data, where numbers appear in diverse and inconsistent formats~\cite{thawani-etal-2021-representing,thawani2021numeracy}. Unlike words, numbers require precise mathematical reasoning, comparison, or context-specific understanding, which standard tokenization and embedding techniques fail to capture adequately. Experiments have shown that representing numbers in a better way can improve NLP model performance~\cite{thawani2021numeracy}. Considering this and the fact that instructions contain many numerical values and significantly impact the semantics of the code, we must emphasize finding a better numeric representation.

One of the significant challenges in handling numeric values within disassembled code is their extensive range. The sheer variety of possible numbers makes it infeasible for any model to learn embeddings for all of them effectively. Previous tools have tackled this issue in different ways. Some entirely removed numeric values from the disassembled code, while others retained smaller numbers and assigned a special token for larger ones. However, these approaches often lack justification for choosing one method over another. Our work addresses this gap by clearly and systematically comparing different approaches, offering valuable insights into their relative performance and effectiveness.

We are comparing two different variations of representing numeric values in disassembled code. They are:

\subsubsection{Default}
This approach represents the baseline method, in which the disassembled code, including its numeric values, is used without modification. This unaltered input serves as a reference point for evaluating and comparing the performance of alternative preprocessing methods.

\subsubsection{Address to Sequential Identifiers \& Hexadecimal Numeric Values to Decimal}
This preprocessing method replaces memory addresses in the code with sequential identifiers and converts all hexadecimal numeric values into their corresponding decimal representations. Large and widely varying memory addresses in disassembled code are highly specific to individual programs or runtime environments, while the representation of numeric values in hexadecimal adds further complexity. These variations make memory addresses and hexadecimal values challenging for tokenizers to process and limit their semantic usefulness for downstream tasks.

The proposed preprocessing method replaces every distinct memory address in the code with a sequential identifier. For example, if the code contains addresses such as 0x1FF0, 0x1FF4, 0x2000, and 0x2AB8, they will be converted into tokens like addr1, addr2, addr3, and addr4. Each distinct address is assigned a unique token, preserving its identity while normalizing its representation into a manageable vocabulary. Similarly, all hexadecimal numeric values in the code are replaced with corresponding decimal representations, ensuring uniformity in numeric formats.
Suppose the vocabulary size is set to 3000. The most frequent tokens in the code, such as mnemonics and other operational strings, will appear multiple times and naturally occupy slots in the vocabulary. Frequently occurring small numbers will also be included in the vocabulary. In contrast, less frequent large numbers are unlikely to appear in the vocabulary and will instead be replaced with a special token, e.g., $<$OOV$>$.
The intuition behind this approach is to normalize memory addresses using unique identifiers, retain frequently used smaller numbers, and eliminate less frequent outliers. This strategy balances the need for effective representation with the constraints of a fixed vocabulary size. An example illustrating the address to sequential identifiers method is shown in Figure~\ref{fig:address_to_words}.

Two memory addresses appear in lines 13, 19, and 25. Additionally, other numeric values are used across different lines. After applying the preprocessing method, the exact values of the addresses are replaced with sequential identifiers, addr1 and addr2. This transformation retains all relevant information. We can still identify these as addresses and understand that lines 12 and 25 call the same function while line 19 calls a different one. The precise values of the addresses are irrelevant for analysis.

Similarly, smaller numbers remain unchanged, while the large numeric value in line 7 is replaced with an OOV token due to its rarity. This replacement highlights its status as an outlier that does not occur frequently. Normalizing memory addresses and handling numeric values appropriately adds value by making the code more structured and interpretable, potentially simplifying the model’s task in downstream applications.

\begin{figure}[h!]
    \centering
    \includegraphics[width=0.4\textwidth]{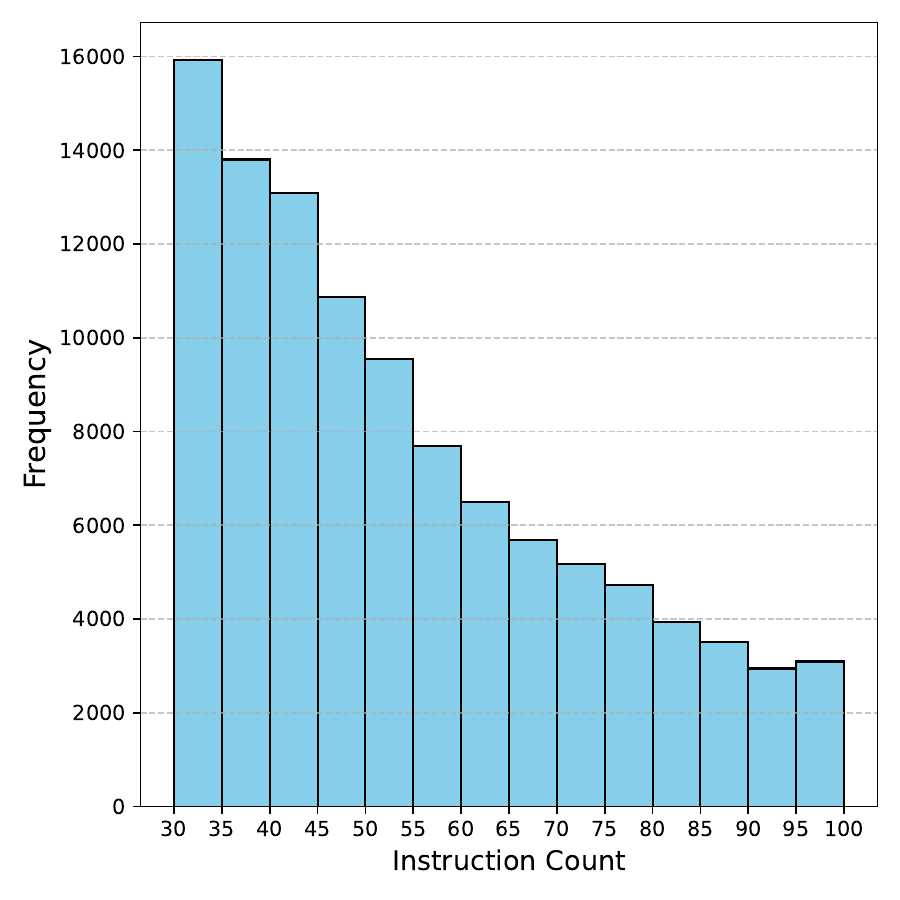} 
    \caption{Frequency distribution of disassembled functions based on the number of instructions per function.}
    \label{fig:histogram}
\end{figure}

\subsection{Dataset}
We scraped code from publicly available GitHub repositories containing C source code to create our dataset. The goal was to ensure a diverse and representative collection of C programs. After collecting the source code, we compiled the programs using the GCC compiler with optimization level 2 and debugging information to preserve metadata. After compilation, we disassembled the binaries using Ghidra~\cite{eagle2020ghidra} to extract individual functions and their signatures. We employed TLSH~\cite{oliver2013tlsh}, a fuzzy hashing technique, to identify and remove duplicate functions to ensure uniqueness and eliminate redundancy in the dataset. Following deduplication, we randomly selected 100,000 functions to form the final dataset. Although the initial pool of functions was significantly more extensive, we opted for this subset due to resource constraints, as our experiments involved extensive computational demands. We filtered the functions in the dataset to include only those containing at least 30 and at most 100 instructions. A histogram of the Frequency distribution of disassembled functions based on the number of instructions per function is depicted in Figure \ref{fig:histogram}. This range was chosen to capture meaningful functionality while avoiding excessively long functions. The resulting dataset provides a robust basis for experimentation, combining diversity from the initial scraping process, uniqueness ensured by deduplication, and a controlled size and complexity range for efficient analysis. We will make our dataset publicly available upon publication.

\subsection{Models}
To assess the impact of the trained tokenizers on downstream model performance, we fine-tuned three models: Llama 3.2 1B, a decoder-only model with causal language modeling (CLM) training objective, the BERT, an encoder-only transformer model, and the BART, encoder-decoder transformer model. The trained tokenizers were used to fine-tune the models, with evaluations conducted on their respective versions of the default and preprocessing disassembly datasets. This setup allowed for a detailed analysis of how tokenization algorithms, vocabulary sizes, and dataset preprocessing influence the models' downstream task performance. The models' configurations are described in \hyperref[Model Architecture and Hyper-Parameters]{Appendix B}.

\subsection{Evaluation}
Our study was structured into two key phases to evaluate the impact of tokenization strategies on downstream model performance: \textbf{intrinsic evaluation}, and \textbf{extrinsic evaluation}.

\subsubsection{Intrinsic evaluation overview}
focused on analyzing tokenizer performance independently of the models, particularly emphasizing the fertility metric, the overlap between the vocabulary generated by different tokenizers, and testing all tokenizers against the out-of-vocabulary issue. The evaluation was performed using a held-out set of 20,000 disassembled functions, ensuring the evaluation data was not used during tokenizer training. This phase aimed to provide insights into the tokenizers' properties without considering their direct impact on model performance.

\textbf{Fertility}, a widely used metric in NLP Scao et al.~\cite{scao2022bloom}; Stollenwerk~\cite{stollenwerk2023training}; Rust et al.~\cite{rust-etal-2021-good}, measures the average number of tokens required to represent a word or document. In our study, we adapted this metric to the domain of \textbf{functions' disassembly code}, where a function's disassembly serves as an analogous unit to a document in NLP. Specifically, fertility in this context quantifies the average number of tokens required to represent the instructions and operands of a disassembled function. This adaptation allows us to assess the compression efficiency and granularity of the tokenizers when applied to assembly code, which, like natural language, contains structural and semantic patterns.

To calculate fertility, we divided the total number of tokens generated by a tokenizer for a dataset of disassembled functions by the total number of instructions in those functions. Instructions were identified using a standardized parsing process that splits assembly code at line breaks. A higher fertility value indicates lower compression efficiency, suggesting that the tokenizer produces more tokens per instruction, which can impact the downstream processing of binary code.

By applying the fertility metric to functions' disassembly code, we gained critical insights into how tokenizers such as BPE, Unigram, and WordPiece with varying vocabulary sizes capture low-level code's structural and semantic information. These insights laid the groundwork for the extrinsic evaluations, where we examined the impact of these tokenizers on the downstream performance of the models.

\subsubsection{Extrinsic evaluation overview}
Extrinsic evaluation assesses the performance of models on downstream tasks to understand the impact of different tokenizers on their effectiveness. In this study, we evaluate the BERT model on masked token prediction accuracy, which aligns with its masked language modeling (MLM) training objective. For the Llama 3.2 model, we evaluate its ability to recover entire disassembled functions, including instructions or parts of instructions, after masking randomly selected tokens. The masked regions are determined by the tokenization strategy employed during the experiments. Although masked token prediction is traditionally a pre-training objective, we included it in our evaluation to measure the models' understanding of disassembly code structure, semantics, and syntax. This analysis underscores the models' ability to interpret low-level assembly instructions and accurately reconstruct missing or obscured tokens.

For the Llama 3.2 model, recovering the entire function disassembly, rather than just the masked tokens, is crucial for assessing its ability to generate coherent and accurate outputs for real-world applications such as code completion, reverse engineering, and vulnerability detection. This evaluation highlights the model's capacity to reconstruct the full execution logic of functions, providing insights into its generative capabilities and robustness in binary program analysis tasks.

Additionally, the Llama 3.2 and BART models were evaluated on the function signature prediction task, a critical downstream task in binary analysis. Function signature prediction involves inferring high-level function prototypes (parameter and return types) from low-level code. This task is important for recovering meaningful symbolic information from stripped binaries, enabling better code comprehension and facilitating downstream tasks such as debugging, optimization, and malware analysis. These evaluations collectively provide a comprehensive view of the tokenizers' impact on model performance across tasks requiring generation and understanding capabilities.

The decision to select the BART-Base model instead of the BERT model to evaluate the function signature prediction downstream task performance was because the BERT is not a generative model and is thus unsuitable for this task. Our choice of models aimed to include a very large model (Llama 3.2) and a smaller model (BART-Base) to provide a comparative perspective. The BART-Base model, being a smaller generative model, offers valuable insights into how model size impacts performance on function signature prediction compared to a larger model like Llama 3.2.

\section{Intrinsic Evaluation of Tokenizers}
We begin by analyzing the fertility scores of the trained tokenizers using an unseen dataset consisting of 20,000 disassembled functions and then examine their vocabulary's overlapping insights.

\subsection{Analysis of Fertility Scores}
\begin{figure*}[!t]
\centering
\subfloat[Default disassembly dataset]{\includegraphics[width=3.5in]{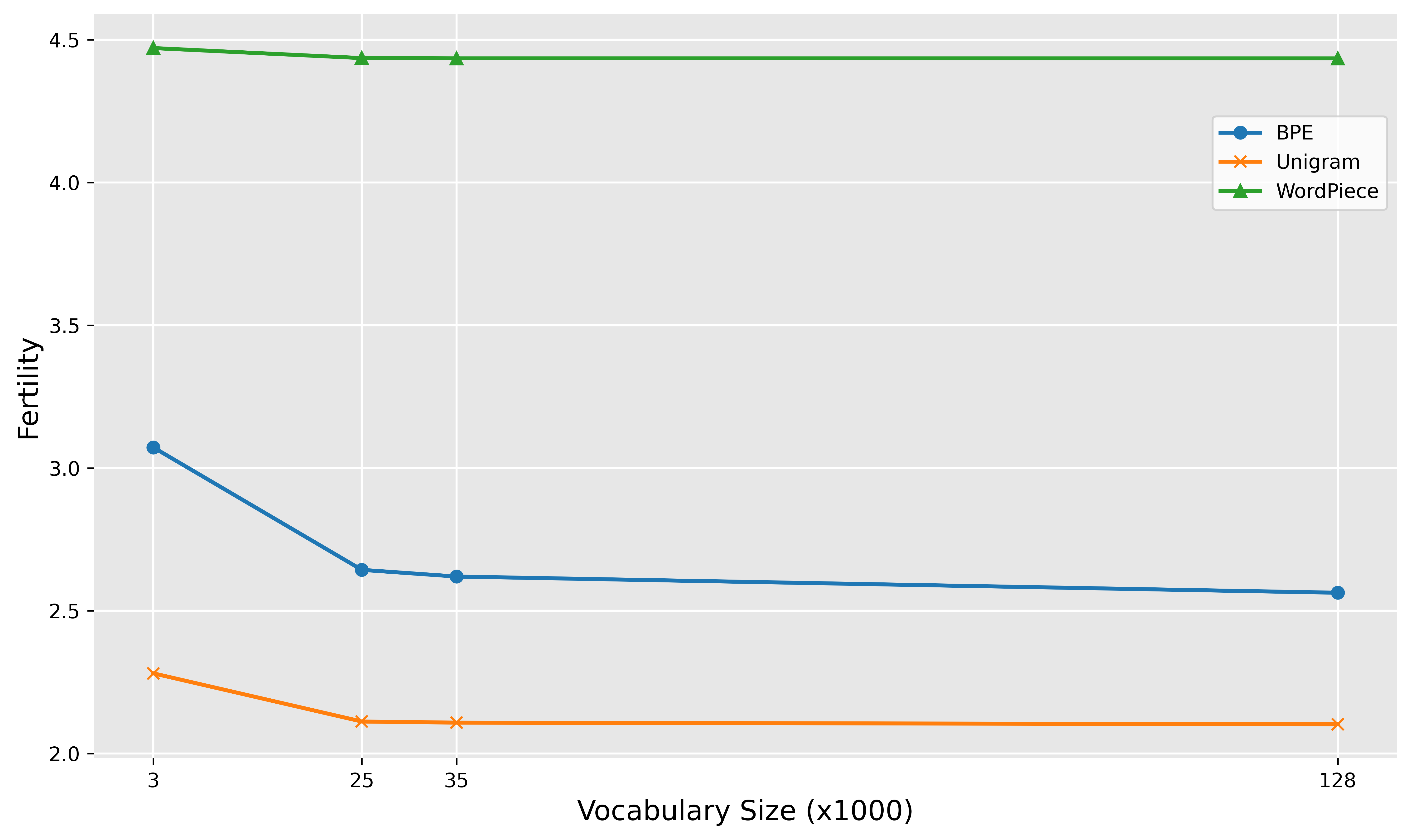}
\label{fig_default_disassembly}}
\hfil
\subfloat[Preprocessed disassembly dataset]{\includegraphics[width=3.5in]{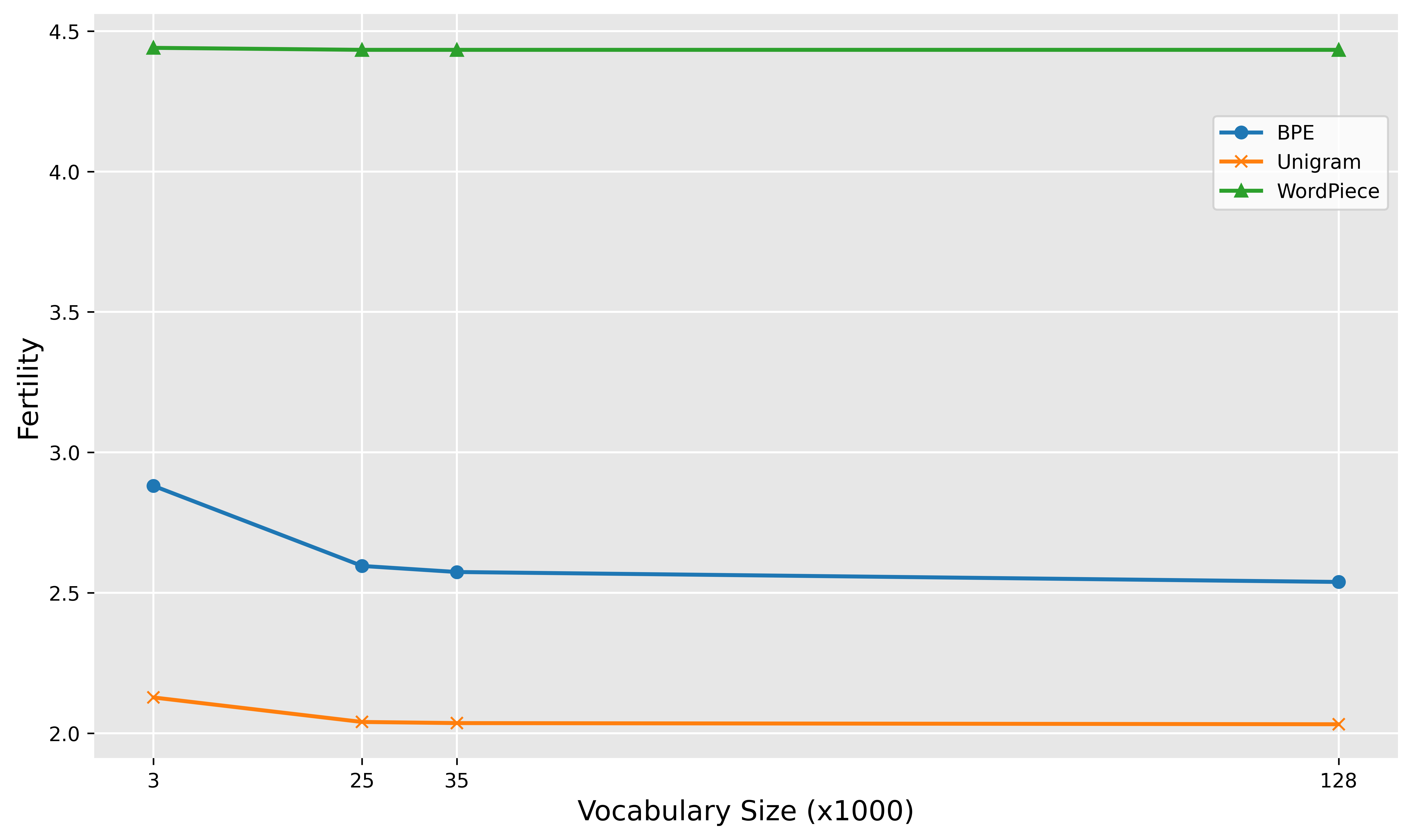}
\label{fig_pre-processed_disassembly}}
\caption{Fertility evaluation comparison between BPE, Unigram, and WordPiece tokenizers on (a) The default disassembly dataset and (b) The Preprocessed disassembly dataset.}
\label{fig_fertility}
\end{figure*}

The fertility study, as described, evaluates the number of tokens BPE, Unigram, and WordPiece tokenizers require to represent instructions in the unseen default and preprocessed disassembly datasets. Fertility measures each tokenizer's efficiency and compression capability. Our observations, based on the fertility score comparison depicted in Figures~\ref{fig_default_disassembly} and \ref{fig_pre-processed_disassembly}, are as follows:

\subsubsection{Default Disassembly Dataset}
\begin{itemize}
    \item \textbf{WordPiece} consistently shows the highest fertility score across all vocabulary sizes (4.5 tokens per instruction), indicating it produces the most tokens per instruction and demonstrates the lowest compression efficiency.
    \item \textbf{Unigram} achieves the lowest fertility score, consistently around 2.0 tokens per instruction, showcasing the highest compression capability among the three tokenizers.
    \item \textbf{BPE} lies between WordPiece and Unigram, with fertility scores decreasing from approximately 3.0 to 2.5 as the vocabulary size increases.
\end{itemize}

\subsubsection{Preprocessed Disassembly Dataset}
\begin{itemize}
    \item Similar trends are observed, where \textbf{WordPiece} maintains the highest fertility score, followed by BPE and Unigram.
    \item o	The preprocessing step slightly reduces fertility for all tokenizers, suggesting better alignment between the tokenizers and the structure of preprocessed disassembly.
\end{itemize}

We conclude that the \textbf{Unigram} is the most efficient tokenizer in terms of compression for both datasets, requiring fewer tokens per instruction, making it ideal for tasks prioritizing compact representations. \textbf{WordPiece}, due to its high fertility, may preserve more granularity in tokenization, which could benefit specific tasks requiring detailed token-level information but at the cost of efficiency. The \textbf{BPE} tokenizer balances compression and granularity, making it a versatile choice for disassembly tasks.

\subsection{Vocabulary Overlap Study}
The vocabulary overlap study examines the percentage of shared vocabularies across BPE, Unigram, and WordPiece tokenizers for four vocabulary sizes (3K, 25K, 35K, 128K). We measure the overlap for both the default and preprocessed disassembly datasets. Our observations, based on vocabulary overlap percentage shown in Table \ref{table_vocab_overlap}, are as follows:

\subsubsection{Overlap Trends}
\begin{itemize}
    \item The vocabulary overlap percentage decreases as the vocabulary size increases, indicating less agreement between tokenizers with larger vocabularies.
    \item \textbf{For the default disassembly dataset:} At a vocabulary size of 3K, the overlap percentage is relatively low (0.75 or 63 tokens), which diminishes as the vocabulary size increases to 128K (0.09 or 187 tokens).
    \item \textbf{For the preprocessed disassembly dataset:} The overlap is slightly higher than in the default dataset, especially at smaller vocabulary sizes (1.04 or 86 tokens at 3K).
\end{itemize}

\subsubsection{Impact of Preprocessing}
Preprocessing enhances vocabulary alignment across tokenizers, as reflected in the higher overlap values for all vocabulary sizes.

We conclude that the overlap between vocabularies is minimal, suggesting that each tokenizer captures unique aspects of the data and may tokenize differently based on its underlying algorithm. Preprocessing the dataset enhances tokenization pattern alignment across the tokenizers, slightly increasing the shared vocabulary. Tasks requiring consistency across tokenizers may benefit from preprocessing to improve uniformity, although the distinct tokenization mechanisms (subword segmentation strategies) will continue to produce unique vocabularies.

In \hyperref[Intrinsic Evaluation of Tokenizers]{Appendix C}, we discuss vocabulary overlap heatmaps, which are shown in Figure~\ref{fig_vocabulary_overlap}. The heatmaps depict the similarity level between different tokenizers by comparing the vocabulary overlapping percentage of different tokenizers across the two datasets (default and preprocessed) and varying vocabulary sizes (3K, 25K, 35K, and 128K).

\subsection{Out-of-Vocabulary Analysis}
All tokenizers (BPE, Unigram, WordPiece) with various vocabulary sizes were evaluated on the test dataset for OOV issues. None of the tokenizers exhibited the OOV problem; they successfully recognized all tokens within their respective vocabulary lists, and no tokens were classified as unk\_token.

\begin{table}[!t]
\renewcommand{\arraystretch}{1.3}
\caption{Vocabulary Overlap Percentage Across Tokenizers for Different Vocabulary Sizes}
\label{table_vocab_overlap}
\centering
\begin{tabular}{|c|c|c|c|c|}
\hline
Vocabulary- & \multicolumn{2}{|c|}{\textbf{Default Disassembly}} & \multicolumn{2}{|c|}{\textbf{Preprocessed Disassembly}} \\
\cline{2-5}
Size & Percentage & \# Tokens & Percentage & \# Tokens \\
\hline
\textbf{3K} & 0.75\% & 63 & 1.04\% & 86 \\
% \hline
\textbf{25K} & 0.13\% & 92 & 0.25\% & 174 \\
% \hline
\textbf{35K} & 0.1\% & 102 & 0.31\% & 267 \\
% \hline
\textbf{128K} & 0.09\% & 187 & 0.44\% & 845 \\
\hline
\end{tabular}
\end{table}

\begin{table}[t!]
\renewcommand{\arraystretch}{1.3}

\caption{
 \textbf{average accuracy} of Masked Token Prediction on the Default and Preprocessed datasets for BERT Across Tokenizers and Vocabulary Sizes}
 
\label{tab:mlm}

\centering
\begin{tabular}{ |p{2cm}|p{1cm}|p{1cm} |p{1cm} |p{1cm} |}
\hline
\multirow{3}{*}{Tokenizer} & \multicolumn{2}{|p{1.8cm}|}
{\textbf{Default \hspace{0.4cm} Disassembly}} & \multicolumn{2}{|p{1.8cm}|}{\textbf{Preprocessed Disassembly}} \\
\cline{2-5}
& Llama- & BERT- & Llama- & BERT- \\
& Accuracy & Accuracy  & Accuracy & Accuracy \\
\hline

\textbf{Unigram-3k}  &70.20  & 73.98 &71.93& 74.25 \\
\textbf{Unigram-25k} &71.65  & 82.86 &73.64& 83.47 \\
\textbf{Unigram-35k} &71.69 & 83.24 &73.62& 84.54 \\

\hline

\textbf{WordPiece-3k}  & 71.20  & 73.38 & 70.92& 75.73 \\
\textbf{WordPiece-25k}  &71.00  & 82.17 & 71.23& 84.45 \\
\textbf{WordPiece-35k}   & 71.67& 83.45 &72.40& 86.49 \\

\hline

\textbf{BPE-3k}  & 70.90 & 72.94 & 72.32& 75.28 \\
\textbf{BPE-25k} & 70.58 & 82.84 & 72.86& 85.65 \\
\textbf{BPE-35k} & 71.60 & 84.28 &72.86& \textbf{86.58} \\

\hline
\textbf{Model-default} &80.48& 85.37&82.30 & 78.08 \\
\hline

\end{tabular}

\end{table}

\begin{table}[!t]
\renewcommand{\arraystretch}{1.3}
\caption{\textbf{Average Accuracy} of Function Parameter and Return Type Prediction on the Default and Preprocessed Datasets for Llama 3.2 and BART Across Tokenizers and Vocabulary Sizes}
\label{table_func_sig_pred}
\centering
\begin{tabular}{ |p{2cm}|p{1cm}|p{1cm} |p{1cm} |p{1cm} |}
\hline
\multirow{3}{*}{Tokenizer} & \multicolumn{2}{|p{1.8cm}|}
{\textbf{Default \hspace{0.4cm} Disassembly}} & \multicolumn{2}{|p{1.8cm}|}{\textbf{Preprocessed Disassembly}} \\
\cline{2-5}
& Llama- & BART- & Llama- & BART- \\
& Accuracy & Accuracy  & Accuracy & Accuracy \\
\hline
\textbf{BPE-3K} & 82.44 & 85.19 & 81.99 & 87.48 \\
\textbf{BPE-25K} & 85.42 & 86.42 & 85.45 & 87.62 \\
\textbf{BPE-35K} & \textbf{85.76} & 86.94 & 85.25 & 87.18 \\
\textbf{BPE-128K} & 85.23 & - & 84.56 & - \\
\hline
\textbf{Unigram-3K} & 74.78 & 84.97 & 75.58 & \textbf{88.81} \\
\textbf{Unigram-25K} & 77.66 & 80.57 & 78.12 & 86.03 \\
\textbf{Unigram-35K} & 77.74 & 66.36 & 77.96 & 81.40 \\
\textbf{Unigram-128K} & 76.88 & - & 77.95 & - \\
\hline
\textbf{WordPiece-3K} & 82.40 & 87.44 & 80.40 & 84.97 \\
\textbf{WordPiece-25K} & 83.25 & 87.53 & 81.75 & 86.48 \\
\textbf{WordPiece-35K} & 76.31 & 87.01 & 83.48 & 87.35 \\
\textbf{WordPiece-128K} & 83.66 & - & 82.04 & -\\
\hline
\textbf{Model-default} & 84.87 & 86.92 & 85.71 & 87.16 \\
\hline

\end{tabular}
\end{table}

\section{Extrinsic Evaluation of Tokenizers}
This section provides an in-depth analysis of how various tokenization algorithms, coupled with different vocabulary sizes and dataset representations, influence performance in two key areas: masked token prediction and the downstream task of function signature prediction. We applied a masking strategy where 15\% of the tokens in each disassembled function were randomly selected and replaced with a special [MASK] token. In contrast, the remaining 85\% of the tokens were left unmasked. This approach ensures a consistent proportion of masked tokens across all functions.

The evaluation aims to uncover the interplay between tokenization strategies and model effectiveness in accurately understanding and generating disassembly code. The experimental results for the Llama 3.2 model are discussed in subsections B and C below, while the experimental results for the BERT and BART models are discussed in subsections A and D, respectively below.

\subsection{Performance Evaluation of Masked Token Prediction with BERT}
In this experiment, we evaluated the performance of several tokenizers in the context of masked token prediction accuracy. This task involves predicting the original token based on its context within a sequence where specific tokens have been replaced with a mask placeholder. We focused solely on the accuracy of predicting the masked tokens, which comprised 15\% of the input sequence, rather than including unmasked tokens in the evaluation. This approach assesses BERT's ability to accurately infer the masked tokens from the surrounding context.

BERT performed better than Llama in most of the experiments. BERT excels in masked token prediction mainly due to its bidirectional context processing. This allows BERT to effectively understand and use the context around masked tokens. Besides, BERT was originally designed to be pre-trained in masked token prediction, which makes it a better fit for this task. 

The evaluation scores presented in Table ~\ref{tab:mlm} show several trends. The model performed better with higher vocabulary-sized tokenizers. The average accuracy increases with higher vocabulary size across all the different tokenizers with both the default and preprocessed datasets.

Likely, the reason for that trend is that with a larger vocabulary size, there are fewer or even no OOV tokens, like in our case. Machine code has a high frequency of numerical values, and the range of the numerical values can be very wide. If the vocabulary size is large, the tokenizers can recognize more outliers, assisting in a better token prediction.

An additional improvement is evident when using the preprocessed dataset compared to the default dataset. This outcome is expected, as the preprocessed dataset is normalized by replacing all addresses with sequential identifiers. Address values in the default dataset can vary widely, introducing significant outliers. Normalizing these values reduces variability and eliminates potential outliers, improving model performance. Among the tokenization algorithms, BPE performed best with larger vocabulary sizes.

Notably, the average masked token prediction accuracy across all tokenizers paired with the preprocessed dataset is consistently higher than with the default dataset. Interestingly, while the BERT default tokenizer performed well with the default dataset, it showed significantly poorer performance when paired with the preprocessed dataset, emphasizing the need to align tokenization strategies with the preprocessing approach.

\subsection{Performance Evaluation of Masked Token Prediction with Llama 3.2}

The evaluation scores can be found in Table ~\ref{tab:mlm}. For masked token prediction, Llama did not perform similarly to BERT, which is expected. BERT's specific design and training make it particularly strong in this area. That's why even with a simpler and lighter design, BERT performed better than Llama. However, the highest accuracy is not the goal of this experiment. The impact of the variations in the tokenizing algorithm, vocabulary size, and preprocessing is much more interesting. The preprocessing algorithm consistently enabled higher performance across different tokenizers and vocabulary sizes. The Llama’s default pre-trained tokenizer, trained with the dataset, showed the highest performance, 80.48\%, and 82.30\% accuracy for the default and preprocessed dataset, respectively. For the custom tokenizers, the highest accuracy we obtained is 71.69\% for the Unigram-35k tokenizer on the default disassembly. For the preprocessed dataset, the highest accuracy observed is 73.64\%.

\subsection{Performance Evaluation of Function Parameters and Return Types Prediction with Llama 3.2}
The Llama 3.2 model evaluation results presented in Table ~\ref{table_func_sig_pred} focus on function signature prediction, a downstream task of predicting function parameters and return types from the function's disassembly:

\subsubsection{Impact of Vocabulary Size}
Table ~\ref{table_func_sig_pred} shows that vocabulary size can marginally impact function signature prediction average accuracy. Increasing the vocabulary size improves accuracy for all tokenizers across the small and moderate vocabulary sizes 3K to 35K (e.g., BPE improves accuracy from 82.44\% for 3K to 85.76\% for 35K on the default dataset and from 81.99\% for 3K to 85.45\% for 25K on the preprocessed dataset).

However, the impact of vocabulary size is less pronounced for WordPiece, where improvements are relatively marginal (e.g., WordPiece improves accuracy from 82.40\% for 3K to 83.25\% for 25K on the default dataset).

\subsubsection{Preprocessed vs. Default Datasets}
On average, the preprocessed dataset improves the default dataset in function signature prediction accuracy. Preprocessing likely enhances the disassembly functions' structural uniformity and semantic clarity, leading to more accurate parameter and return type predictions.

\subsubsection{Impact of Tokenization Algorithms}
The performance of the tokenization algorithms differs across datasets and vocabulary sizes:

\begin{itemize}
    \item \textbf{BPE Tokenizer:} Consistently achieves the highest average accuracy across all vocabulary sizes for the default and preprocessed datasets (e.g., 85.76\% for BPE-35K on the default dataset).
    \item \textbf{Unigram Tokenizer:} This tokenizer shows the lowest average accuracy among all tokenizers, with results generally below 80\% across vocabulary sizes (e.g., 74.78\% for Unigram-3K on the default dataset and 75.58\% for Unigram-3K on the preprocessed dataset).
    \item \textbf{WorPiece Tokenizer:} Performs moderately well, with accuracy slightly behind BPE but significantly better than Unigram (e.g., 83.66\% for WordPiece-128K on the default dataset).
\end{itemize}

The Llama 3.2 model's default pre-trained tokenizer outperformed the Unigram tokenizer on both datasets and all vocabulary sizes, and on average, it achieved slightly higher accuracy than the WordPiece tokenizer across both datasets and all vocabulary sizes. However, the BPE tokenizer achieved very close average accuracy to the Llama 3.2 pre-trained tokenizer, particularly on vocabulary sizes 25K to 128K.

The BPE achieved an average accuracy of 85.76\% for 35K vocabulary size on the default dataset, slightly higher than the average accuracy of the Llama 3.2 model’s default pre-trained tokenizer on both datasets.

\subsection{Performance Evaluation of Function Parameters and Return Types Prediction with BART}
The evaluation scores of the signature prediction task with BART are presented in Table ~\ref{table_func_sig_pred}.
BART performed best with the smallest vocabulary size and preprocessed disassembly for function signature prediction. Across the tokenizers, the performance is mixed. For the default disassembly, WordPiece performed well consistently across all the vocabulary sizes. With preprocessed disassembly,  Unigram-3k performed best with 88.81\% accuracy. Similarly, BPE also performed well with smaller vocabulary sizes. This means a combination of preprocessing and smaller vocabulary size represents the token in a better way for the model to understand the parameters and return type work in a function. The preprocessing step normalizes the addresses, which is beneficial for the signature prediction task because the specific values of the addresses are irrelevant for this particular task. It is enough to know that a token is an address, as the exact value is irrelevant.

\section{Discussion}
The choice of tokenization algorithm, vocabulary size, and dataset representation is crucial and should align with the model type and task. For instance, in the masked token prediction pre-training task, the BERT model paired with a BPE tokenizer and a moderate vocabulary size of 35K, applied to a preprocessed machine code dataset, emerged as the optimal configuration among the evaluated options.

Similarly, for function signature prediction, the BART model paired with a Unigram tokenizer with a small vocabulary size of 3K and preprocessed machine code demonstrated the best performance.

A notable finding is the consistent benefit of dataset preprocessing, which enhances downstream performance, particularly for smaller to moderate vocabulary sizes across all tokenizers and models.

However, the performance gains from preprocessing were negligible for the models' default pre-trained tokenizers. Interestingly, BERT's default tokenizer achieved higher average accuracy on masked token prediction tasks using the default dataset compared to the preprocessed version, underscoring the importance of task-specific dataset-tokenizer alignment.

\textbf{Insights from the Intrinsic Evaluation of Tokenizers:} The intrinsic evaluation highlights key trade-offs between tokenization efficiency, granularity, and alignment. Unigram demonstrates superior compression with the lowest fertility scores, making it ideal for tasks prioritizing compact representations. WordPiece, with higher fertility scores, provides granular tokenization, which may benefit tasks requiring detailed token-level information. BPE balances efficiency and granularity, offering versatility for disassembly tasks.

Preprocessing improves tokenization efficiency and slightly enhances vocabulary alignment across tokenizers. However, minimal overlap among tokenizers, especially with larger vocabularies, suggests that each algorithm captures unique data characteristics. Despite this, preprocessing aids in achieving greater uniformity in tokenization patterns, which could benefit tasks requiring consistency.

\section{Limitations}
Despite the scope of our study, it faces the following limitations:

\subsubsection{Lack of Hyperparameter Optimization}
We did not conduct extensive hyperparameter optimization for each tokenizer to minimize computational time costs and maintain focus on the study's primary objectives. This decision, however, may have constrained the potential performance gains achievable with fine-tuned configurations. Additionally, exploring the impact of hyperparameters such as learning rate, batch size, or dropout rates on downstream tasks could provide valuable insights. Future work could investigate these interactions to identify optimal configurations that enhance the alignment between tokenizers and models across diverse tasks.

\subsubsection{Tokenizer Implementation Variants} Our study relied primarily on specific implementations of tokenizers, such as those provided by the Hugging Face library. While this ensures compatibility with the models used in our study, alternative implementations, such as SentencePiece, may yield different results. Investigating the impact of implementation details on tokenization and downstream performance remains an area for further exploration.

\subsubsection{Intrinsic and Extrinsic Correlation Analysis} We did not study the correlation between the intrinsic properties of tokenizers (e.g., vocabulary size, token overlap, token distribution) and their extrinsic evaluation on downstream task performance. Understanding this relationship could provide deeper insights into how tokenizer design impacts model behavior and performance across tasks, and we encourage future work to explore this dimension.

\subsubsection{Scaling to Larger Models} While our study focused on models with up to 1 billion parameters, we did not evaluate the tokenizers' performance on larger models. Extending the evaluation to larger architectures may uncover additional insights, as tokenization effects could behave differently in models with significantly more parameters.

\subsubsection{Real-world Dataset and Task Coverage} Our evaluation was conducted on specific downstream tasks and machine code datasets. While these tasks and datasets are relevant to the context of this work, they may not fully represent the diversity of real-world applications. Future studies should extend the evaluation to a broader range of datasets and tasks to validate the generalizability of our findings.

By addressing these limitations, future research can refine our understanding of tokenization algorithms, exploring their intrinsic properties, broader applicability, and robustness across diverse tasks, model architectures, and evaluation settings.

\section{Conclusion \& Future Work}
This study on tokenization algorithms for binary code analysis highlights the critical role of tokenization strategies in optimizing the performance of LLMs and transformer-based models. By evaluating the intrinsic properties of various tokenizers and their extrinsic performance on downstream tasks like function signature prediction, we demonstrated that both the choice of tokenization algorithm and vocabulary size significantly influence model outcomes. Although we did not directly study the impact of intrinsic tokenizer properties on downstream task performance, our findings emphasize the importance of selecting tokenization strategies that align with task-specific requirements. Additionally, our experiments underscore the value of preprocessing machine code tailored to the context of the downstream task. The optimal preprocessing strategy, however, is highly task-dependent and requires careful consideration.

Since NLMs were not originally designed for binary analysis tasks, our findings provide valuable insights into how binary code can be effectively represented for such models. This representation step is essential and has a large potential impact on the model's performance. Overall, this study establishes a foundational understanding of selecting appropriate tokenization and preprocessing strategies for leveraging NLMs in binary code analysis tasks.

In future work, we aim to expand our investigation by applying tokenizers to larger datasets that introduce greater structural diversity and dependency varieties in machine code. This will allow us to better understand how tokenization approaches perform with more complex data representations. Additionally, we plan to explore alternative tokenization methodologies, such as comparing the SentencePiece implementation with Hugging Face's implementation, to identify nuances in their impact on model performance.

Furthermore, we intend to scale our evaluations to larger language models exceeding 1 billion parameters, enabling us to assess how tokenizer choices influence performance in more powerful architectures. Finally, we will broaden our evaluation scope by testing tokenizers on a wider range of real-world downstream tasks, ensuring the practical relevance of our findings.

\section*{acknowledgment}

This manuscript has been assigned LA-UR-25-20366. This research was funded by the Nuclear Weapons Cyber Assurance Laboratory (NWCAL) at Los Alamos National Laboratory. The authors gratefully acknowledge the support provided, which made this work possible.

\bibliographystyle{IEEEtran}  % Choose the IEEEtran BibTeX style
\bibliography{bibliography}    % Name of your .bib file (without the .bib extension)

\section*{Appendix}

\subsection{Tokenizer Hyper-Parameters}
\label{Tokenizer Hyper-Parameters}
The \textcolor{blue}{\href{https://huggingface.co/learn/nlp-course/en/chapter6/8}{Hugging Face Tokenizer library}} was the primary tool for configuring the tokenizers' hyperparameters. To evaluate the impact of different tokenization strategies on downstream performance tasks involving disassembled code, we carefully tailored the configurations for each tokenizer, systematically varying the vocabulary sizes, as presented in Table~\ref{table_tokenizer_hyperparameters}. Parameters not listed in Table~\ref{table_tokenizer_hyperparameters} were kept at their default values.

\begin{table}[!t]
\renewcommand{\arraystretch}{1.3}
\caption{Hyperparameter Configurations Options for BPE, Unigram, and WordPiece Tokenizers in this Study}
\label{table_tokenizer_hyperparameters}
\centering
\begin{tabular}{|c|c|c|}
\hline
Tokenizer & Hyper-Parameter & Value(s) \\
\hline
\multirow{13}{*}{\textbf{BPE}} & model\_type & BPE \\
                              & normalization\_rule & NFD, lowercase \\
                              & pre\_tokenizer\_type & ByteLevel \\
                              & add\_prefix\_spac & False\\
                              & use\_regex & False \\
                              & trainer\_type & BpeTrainer \\
                              & \multirow{2}{*}{vocab\_size} & 3K \textbar{} 25K \\
                              &             & 35K \textbar{} 128K \\
                              & post\_processor\_type & ByteLevel \\
                              & trim\_offsets & False \\
                              & decoder\_type & ByteLevel \\
                              & \multirow{2}{*}{special\_tokens} & $<$unk$>$, $<$/s$>$, $<$s$>$, \\
                              &                                  & [PAD], [MASK] \\

\hline
\multirow{14}{*}{\textbf{Unigram}} & model\_type & Unigram \\
                              & normalization\_rule & NFD, lowercase \\
                              & pre\_tokenizer\_type & ByteLevel \\
                              & add\_prefix\_spac & False\\
                              & use\_regex & False \\
                              & trainer\_type & UnigramTrainer \\
                              & \multirow{2}{*}{vocab\_size} & 3K \textbar{} 25K \\
                              &             & 35K \textbar{} 128K \\
                              & post\_processor\_type & ByteLevel \\
                              & trim\_offsets & False \\
                              & decoder\_type & ByteLevel \\
                              & \multirow{3}{*}{special\_tokens} & $<$unk$>$, $<$/s$>$, $<$s$>$, \\
                              &                                  & $<$cls$>$, $<$sep$>$, \\
                              &                                  & [PAD], [MASK] \\

\hline
\multirow{14}{*}{\textbf{WordPiece}} & model\_type & WordPiece \\
                              & normalization\_rule & NFD, lowercase \\
                              & pre\_tokenizer\_type & BertPreTokenizer \\
                              & trainer\_type & WordPieceTrainer \\
                              & \multirow{2}{*}{vocab\_size} & 3K \textbar{} 25K \\
                              &             & 35K \textbar{} 128K \\
                              & decoder\_type & WordPiece \\
                              & prefix & "\#\#" \\
                              & \multirow{3}{*}{special\_tokens} & [UNK], $<$/s$>$, $<$s$>$, \\
                              &                                  & $<$cls$>$, $<$sep$>$, $<$nln$>$, \\
                              &                                  & [PAD], [MASK] \\
\hline
\end{tabular}
\end{table}

\subsection{Model Architecture and Hyper-Parameters}
\label{Model Architecture and Hyper-Parameters}
In this study, we fine-tuned three models with distinct architectures and hyperparameter configurations, as described in the sections below, to comprehensively evaluate the impact of different tokenization strategies on downstream performance. The detailed model architecture and fine-tuning hyperparameters are presented in Table~\ref{table_model_hyperparameters}, providing a clear overview of the experimental parameters.

\subsubsection{Llama 3.2 Decoder-Only Model}
The pre-trained Llama 3.2 model with 1B parameters represents the smallest architecture in the Llama 3 series, designed to provide efficient performance while minimizing computational overhead. As part of the Llama series, which excels in various NLP tasks such as text generation, summarization, and question-answering, the 1B model balances model complexity with resource efficiency. It employs a transformer-based architecture optimized for generative and comprehension tasks, leveraging a robust 128K vocabulary size for precise tokenization and language representation. Despite its smaller size, Llama 3.2 demonstrates impressive capabilities in handling tasks that require understanding complex language structures, making it an ideal choice for resource-constrained environments or domain-specific fine-tuning.

We leveraged the Llama 3.2 model with 1B parameters to fine-tune it for downstream tasks using tokenizers trained under various configurations. Our implementation followed guidelines from the Hugging Face’s training repository \textcolor{blue}{\href{https://github.com/philschmid/deep-learning-pytorch-huggingface/blob/064795ec2a900ffb548099b791284210727f6ba4/training/scripts/run_fsdp_qlora.py}{deep-learning-pytorch-huggingface}}. This setup allowed us to systematically evaluate the impact of tokenization strategies on performance, particularly in tasks requiring the understanding of complex disassembled code structures.

\subsubsection{BERT Encoder-Only Model}
One of the reasons for using a pre-trained BERT-based model was to evaluate how a smaller, encoder-only transformer model performs on binary analysis tasks. However, for some experiments, modifications were necessary to adapt BERT to our specific requirements. The default BERT vocabulary size is 30,522, but we experimented with alternative vocabulary sizes, including 3,000, 25,000, and 35,000, requiring us to train custom tokenizers. Additionally, BERT's default configuration supports a maximum input token length of 512. To accommodate longer input sequences of up to 1,024 tokens, we extended the model's positional embeddings. Hugging Face hosts the specific BERT-based model variant we used in this experiment and can be found \textcolor{blue}{\href{https://huggingface.co/microsoft/MiniLM-L12-H384-uncased}{here}}.

\subsubsection{BART Encoder-Decoder Model}
BART’s model architecture leverages the strengths of both encoder and decoder components in the Transformer model. It is designed for natural language processing tasks that combine bidirectional encoding and autoregressive decoding. One advantage of using BART over BERT is that BART can handle 1024 tokens by default, and no modification was needed for our experiment. We used the \textcolor{blue}{\href{https://huggingface.co/facebook/bart-base}{pre-trained BART-base}} model version, which Hugging Face hosts.

\begin{table*}[!t]
\renewcommand{\arraystretch}{1.3}
\caption{Overview of the Architecture and Hyperparameter Configurations of the Models Used in the Study}
\label{table_model_hyperparameters}
\centering
\begin{tabular}{|c|c|c|}
\hline
Model & Hyper-Parameter & Value(s) \\
\hline
\multirow{14}{*}{\textbf{Llama 3.2}} & Hidden Size & 2048 \\
                              & Number of Attention Heads & 32 \\
                              & Number of Hidden Layers & 16 \\
                              & Context Window & 1024\\
                              & Learning rate & 0.0002 \\
                              & Learning rate scheduler & linear \\
                              & Gradient accumulation steps & 2 \\
                              & Optimizer & adamw\_torch \\
                              & Max. gradient norm. & 0.3 \\
                              & Warmup ratio & 0.03 \\
                              & Precision & tf32 \\
                              % & LoRA-alpha & 8 \\
                              % & LoRA-dropout & 0.05 \\
                              % & LoRA-rank & 32 \\

\hline

                              &Hidden Size & 384\\
                              & Number of Attention Heads & 12 \\
                              & Number of Hidden Layers & 12 \\
                              & Max Position Embeddings& 512\\
                              \textbf{BERT}& Learning rate & 0.0003 \\
                              &Learning rate schedular & linear schedule with warmup \\
                              & Weight decay & 0.0001 \\
                              & Gradient accumulation steps & 1 \\
                              & Optimizer & adamw\_torch \\

\hline
                              &Hidden layers& 6\\
                              & Number of Encoder Attention Heads & 12 \\
                              & Number of Decoder Attention Heads & 12 \\
                              & Number of Hidden Layers & 12 \\
                              & Max Position Embeddings& 1024\\
                              \textbf{BART}&Decoder layers & 6 \\
                              &Encoder layers&6\\
                              & Learning rate & 0.00005 \\
                              &Learning rate decay & Constant \\
                              & Gradient accumulation steps & 1 \\
                              & Optimizer & adamw\_torch \\
\hline
\end{tabular}
\end{table*}

\subsection{Intrinsic Evaluation of Tokenizers}
\label{Intrinsic Evaluation of Tokenizers}
\begin{figure*}[!t]
\centering
\subfloat{\includegraphics[width=3.5in]{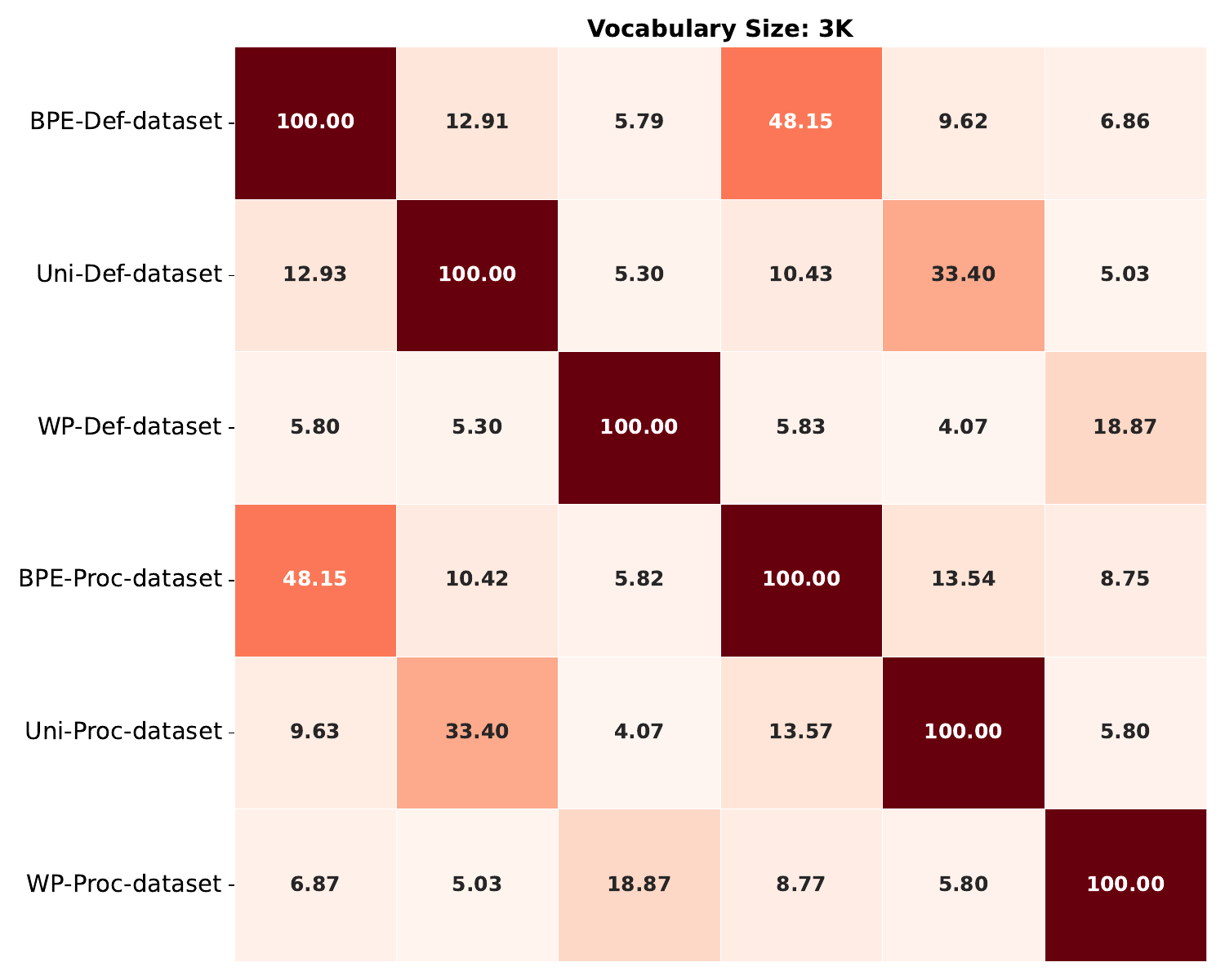}
\label{fig_vocab-3K_overlap}}
\hfil
\subfloat{\includegraphics[width=3.5in]{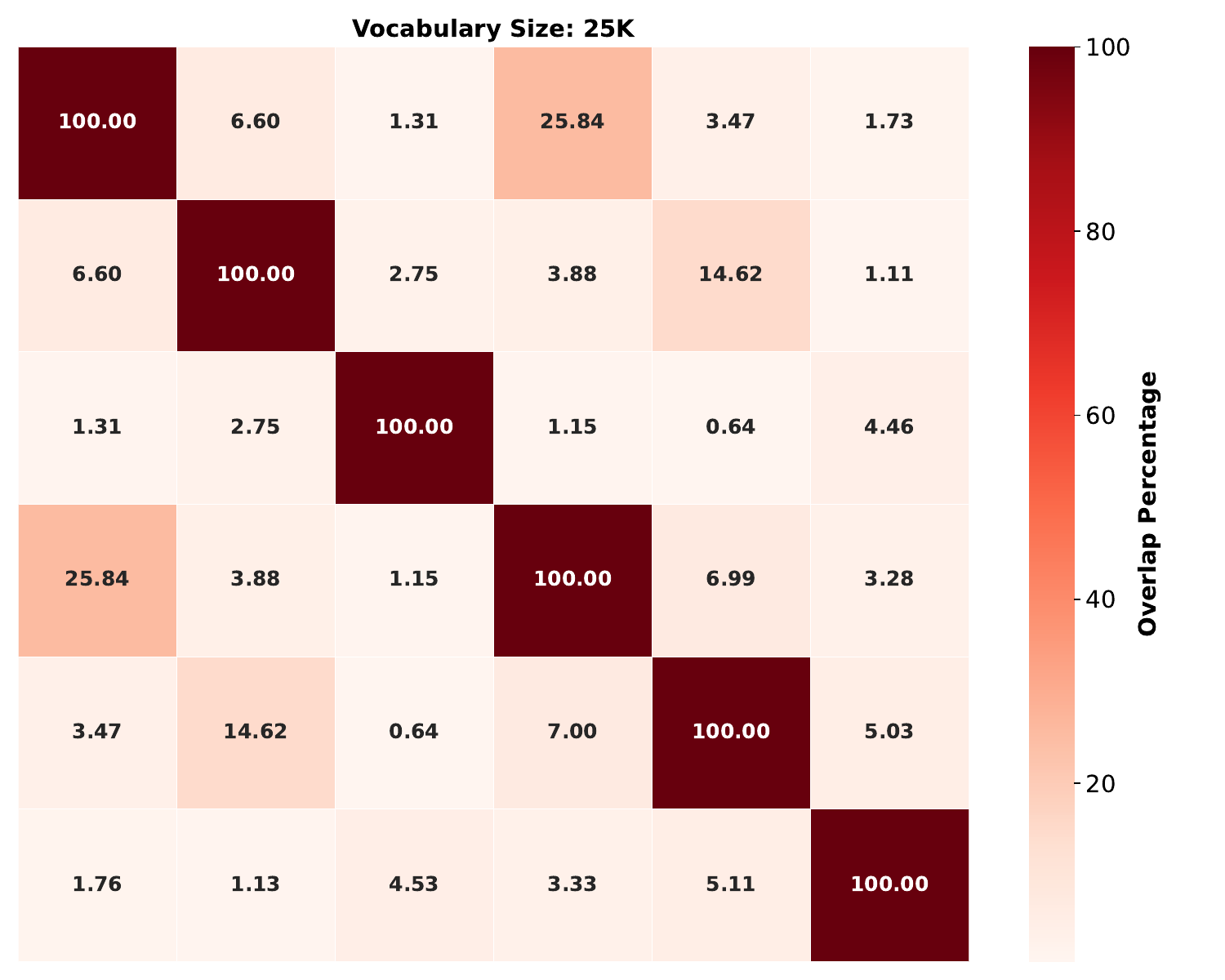}
\label{fig_vocab-25K_overlap}}
\hfil
\subfloat{\includegraphics[width=3.5in]{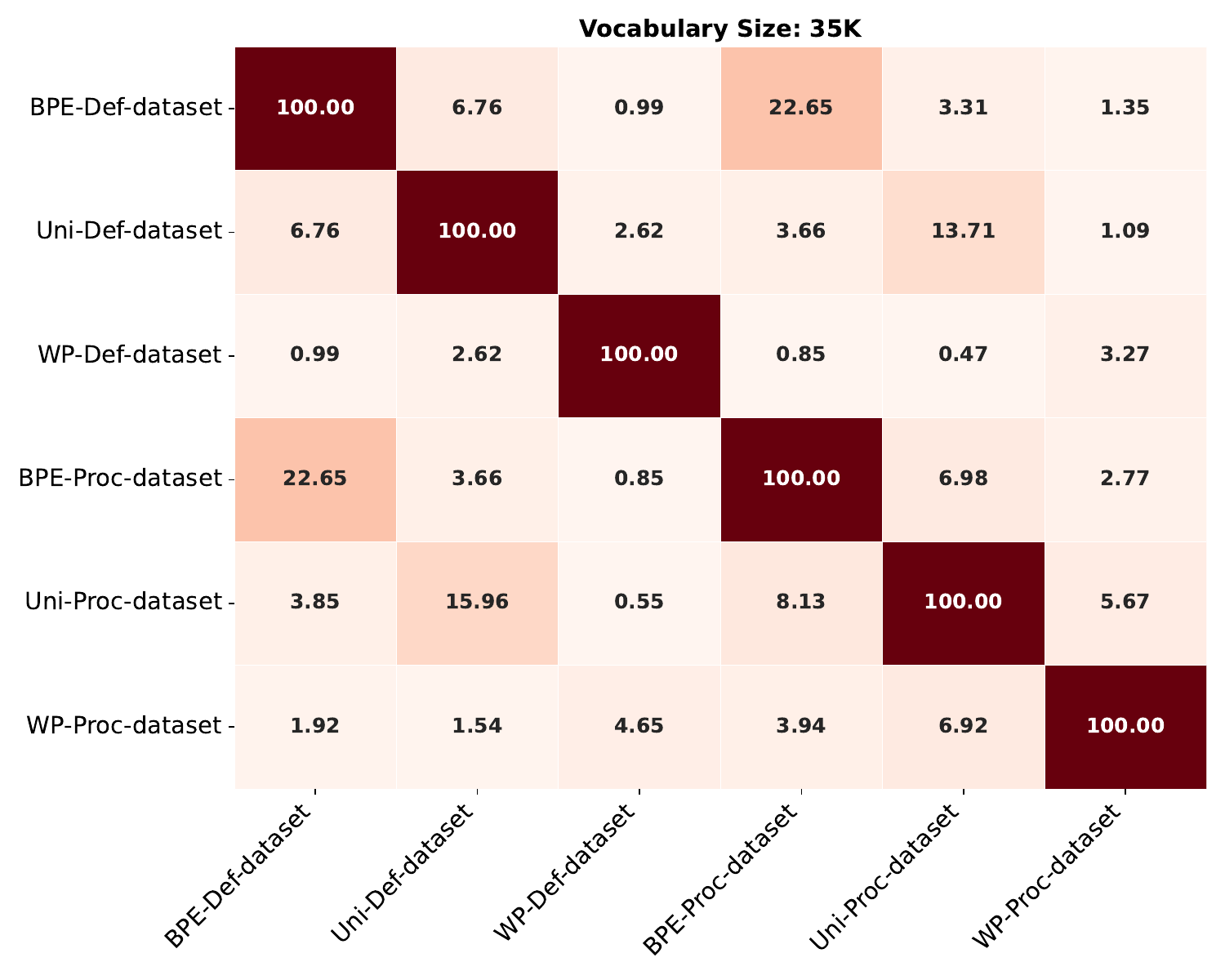}
\label{fig_vocab-35K_overlap}}
\hfil
\subfloat{\includegraphics[width=3.5in]{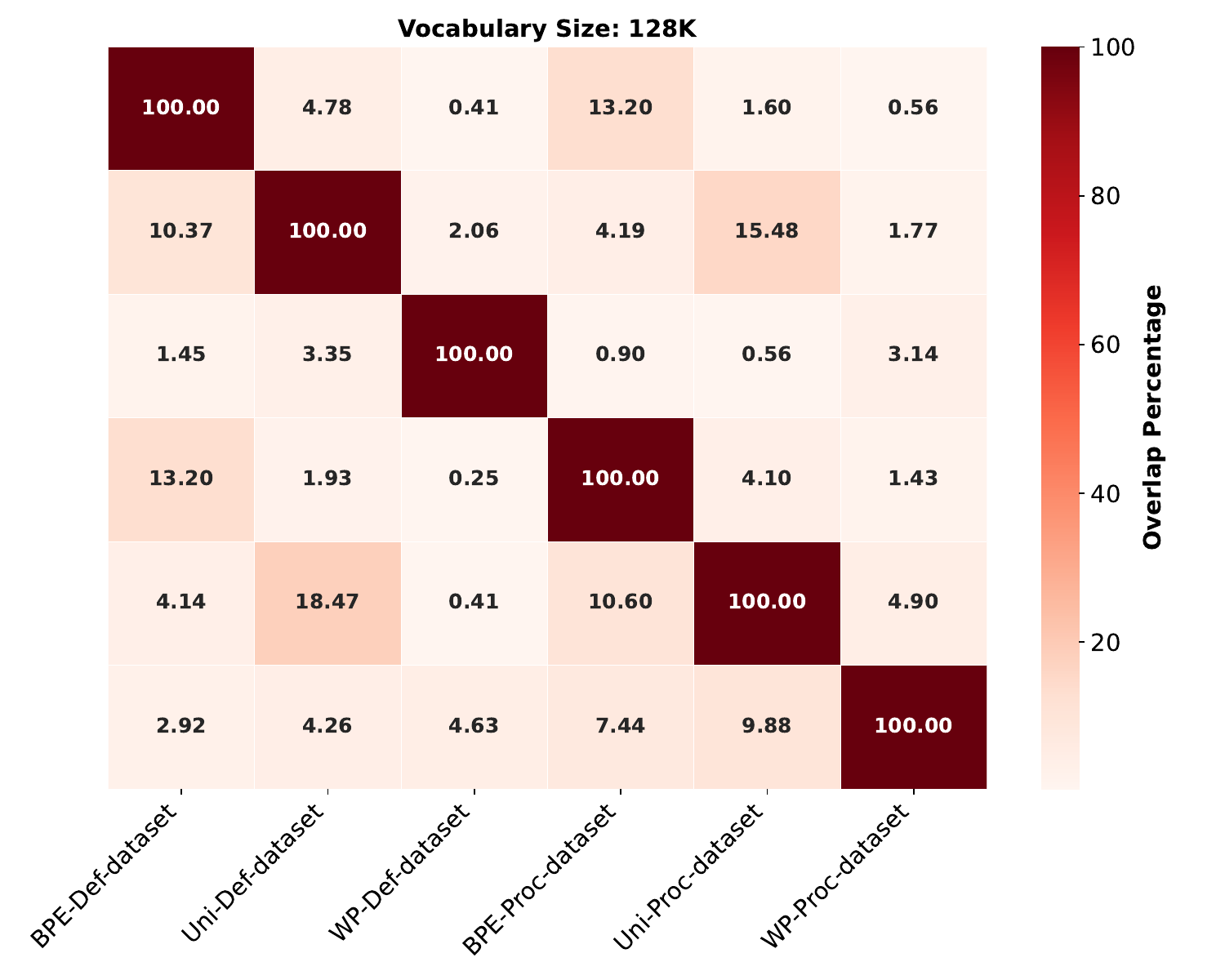}
\label{fig_vocab-128K_overlap}}
\caption{Vocabulary overlap heatmaps for tokenizers across default and preprocessed datasets at four vocabulary sizes: 3K, 25K, 35K, and 128K.}
\label{fig_vocabulary_overlap}
\end{figure*}

In addition to examining vocabulary overlaps among different tokenizers within the same dataset type (either exclusively Default or exclusively Preprocessed) across four vocabulary sizes, we also investigated the vocabulary overlap between tokenizers applied to the Default and Preprocessed datasets as shown in Figure ~\ref{fig_vocabulary_overlap}. This extended analysis provided insights into how preprocessing influences tokenization consistency and vocabulary alignment across datasets and tokenization algorithms.

The vocabulary overlap between tokenizers is notably low, especially across different datasets and tokenization algorithms. The overlap decreases as the vocabulary size increases, highlighting the divergence in subword segmentation strategies used by BPE, Unigram, and WordPiece tokenizers. Preprocessing the dataset moderately improves overlap within the same tokenizer type but does not significantly increase overlap between different tokenizers, demonstrating that tokenization algorithms inherently yield distinct vocabulary sets.

\subsection{Illustrative Examples for the Tokenization Behavior}
\label{Illustrative Examples for the Tokenization Behavior}

Tables~\ref{table_default_disassembly_example} and~\ref{table_pre-processed_disassembly_example} show the tokenization results for five representative disassembled instructions from a binary function presented in two formats:
\\
\textbf{Default disassembly:}\\
\texttt{"ENDBR64\textbackslash nCMP EDI,ESI\textbackslash nJGE 0x000012ce\textbackslash nPUSH R13\textbackslash nMOV R8D,EDI"}
\\
\\
\textbf{Preprocessed disassembly:}\\
\texttt{"ENDBR64\textbackslash nCMP EDI,ESI\textbackslash nJGE addr14\textbackslash nPUSH R13\textbackslash nMOV R8D,EDI"}

The default disassembly retains original numeric values, including memory addresses, while the preprocessed version replaces memory addresses with human-readable sequential identifiers like addr14. The tokenization results were evaluated across three tokenizers (BPE, Unigram, and WordPiece) and the 25K vocabulary size. The tokenization behavior of the default disassembly is presented in Table~\ref{table_default_disassembly_example} while the tokenization behavior of the preprocessed disassembly is presented in Table~\ref{table_pre-processed_disassembly_example}.

Table~\ref{table_pre-processed_disassembly_function_example} shows the tokenization results for an entire disassembled function instructions using the BPE tokenizer with varying vocabulary sizes. The disassembly is presented in the preprocessed format:
\\

\textbf{Preprocessed disassembly:}\\
\texttt{"ENDBR64\textbackslash nPUSH RBP\textbackslash nMOV RBP,RDI\textbackslash nMOV RDI,RSI\textbackslash nPUSH RBX\textbackslash nMOV RBX,RSI\textbackslash nSUB RSP,8\textbackslash nCALL addr0\textbackslash nTEST EAX,EAX\textbackslash nJLE addr5\textbackslash nSUB EAX,1\textbackslash nMOVZX R9D,word ptr [addr8]\textbackslash nMOV RSI,RBX\textbackslash nXOR EDX,EDX\textbackslash nMOVZX R8D,word ptr [addr7]\textbackslash nLEA RDI,[RBX + RAX*1 + 1]\textbackslash nJMP addr3\textbackslash nCMP CL,10\textbackslash nJNZ addr6\textbackslash nMOV word ptr [RAX],R8W\textbackslash nADD EDX,2\textbackslash nADD RSI,1\textbackslash nCMP RDI,RSI\textbackslash nJZ addr4\textbackslash nMOVZX ECX,byte ptr [RSI]\textbackslash nMOVSXD RAX,EDX\textbackslash nADD RAX,RBP\textbackslash nCMP CL,9\textbackslash nJNZ addr1\textbackslash nADD RSI,1\textbackslash nMOV word ptr [RAX],R9W\textbackslash nADD EDX,2\textbackslash nCMP RDI,RSI\textbackslash nJNZ addr3\textbackslash nMOVSXD RDX,EDX\textbackslash nADD RBP,RDX\textbackslash nMOV byte ptr [RBP],0\textbackslash nADD RSP,8\textbackslash nPOP RBX\textbackslash nPOP RBP\textbackslash nRET\textbackslash nMOV byte ptr [RAX],CL\textbackslash nADD EDX,1\textbackslash nJMP addr2\textbackslash n"}

The results in Table~\ref{table_pre-processed_disassembly_function_example} depict an example of the tokenization behavior of the BPE tokenizer with varying vocabulary sizes: 3K, 25K, 35K, and 128K. It is notable that the number of tokens generated decreases as the tokenizer's vocabulary size increases. This behavior is inherent to the design of the BPE algorithm, which creates its vocabulary by iteratively merging pairs of frequently co-occurring sub-tokens in the dataset.
With a smaller vocabulary size, the tokenizer splits words or instructions into smaller sub-tokens to fit within the limited dictionary. As the vocabulary size increases, the BPE algorithm incorporates more frequent sub-token pairs into its dictionary, allowing it to merge smaller sub-tokens into longer, semantically meaningful tokens. Consequently, this leads to fewer overall tokens being generated for the same input text, as larger vocabulary sizes provide more complete token representations. This process highlights the efficiency of the BPE algorithm in balancing granularity and compression based on the frequency of token occurrences in the dataset and the constraints of the vocabulary size. The color-coded example in Table~\ref{table_pre-processed_disassembly_function_example} visually demonstrates how these changes manifest in tokenizing the same disassembled function.

\begin{table*}[!t]
\renewcommand{\arraystretch}{1.3}
\caption{Tokenization Example for Five Default Disassembled Instructions}
\label{table_default_disassembly_example}
\centering
\begin{tabular}{|l|c|c|}
\hline
Tokenizer & Tokens & \# of Tokens\\
\hline

\textbf{BPE-25K} & \texttt{"endbr64\textbackslash n","cmp edi,esi\textbackslash n",\textbf{"jge 0x000012","ce\textbackslash n"},"push r13\textbackslash n","mov r8d,","ed","i"} & \textbf{8} \\
\hline
\multirow{2}{*}{\textbf{Unigram-25K}} & \texttt{"e","nd","br","64","\textbackslash n","cmp edi,","esi","\textbackslash n",\textbf{"jge 0x000012c","e"},"\textbackslash n",} & \multirow{2}{*}{\textbf{16}} \\
                                      & \texttt{"push r1","3","\textbackslash n","mov r8d,","edi"} & \\
\hline
\textbf{WordPiece-25K} & \texttt{"endbr64","cmp","edi",",","esi",\textbf{"jge","0x000012ce"},"push","r13","mov","r8d",",","edi"} & \textbf{13} \\

\hline
\end{tabular}
\end{table*}

\begin{table*}[!t]
\renewcommand{\arraystretch}{1.3}
\caption{Tokenization Example for Five Preprocessed Disassembled Instructions}
\label{table_pre-processed_disassembly_example}
\centering
\begin{tabular}{|l|c|c|}
\hline
Tokenizer & Tokens & \# of Tokens\\
\hline

\textbf{BPE-25K} & \texttt{"endbr64\textbackslash n","cmp edi,esi\textbackslash n",\textbf{"jge addr14\textbackslash n"},"push r13\textbackslash n","mov r8d,","ed","i"} & \textbf{7} \\
\hline
\multirow{2}{*}{\textbf{Unigram-25K}} & \texttt{"e","nd","b","r64","\textbackslash n","cmp edi,","esi","\textbackslash n",\textbf{"jge addr1","4"},"\textbackslash n",} & \multirow{2}{*}{\textbf{16}} \\
                                      & \texttt{"push r1","3","\textbackslash n","mov r8d,","edi"} & \\
\hline
\textbf{WordPiece-25K} & \texttt{"endbr64","cmp","edi",",","esi",\textbf{"jge","addr14"},"push","r13","mov","r8d",",","edi"} & \textbf{13} \\

\hline
\end{tabular}
\end{table*}

\begin{table*}[!t]
\renewcommand{\arraystretch}{1.3}
\caption{Tokenization Example of an Entire Preprocessed Disassembled Function Instructions}
\label{table_pre-processed_disassembly_function_example}
\centering
\begin{tabular}{|l|c|c|}
\hline
Tokenizer & Tokens & \# of Tokens\\
\hline

\multirow{12}{*}{\textbf{BPE-3K}} &\texttt{"endbr64\textbackslash n", "push rbp\textbackslash n", "mov rbp,rdi\textbackslash n", "mov rdi,rsi\textbackslash n", "push rbx\textbackslash n",} & \multirow{12}{*}{\textbf{67}} \\

& \texttt{"mov rbx,rsi\textbackslash n", "sub rsp,8\textbackslash n", "call addr0\textbackslash n", "test eax,eax\textbackslash n",} & \\

& \texttt{"jle addr5\textbackslash n", "sub eax,1\textbackslash n", \textbf{\textcolor{blue}{"movzx r9d,"}}, \textbf{\textcolor{blue}{"word ptr [addr"}}, \textbf{\textcolor{blue}{"8]\textbackslash n"}},} & \\

& \texttt{"mov rsi,rbx\textbackslash n", "xor edx,edx\textbackslash n", \textbf{\textcolor{red}{"movzx"}}, \textbf{\textcolor{red}{" r8d,"}}, \textbf{\textcolor{red}{"word ptr [addr"}}, \textbf{\textcolor{red}{"7]\textbackslash n"}},} & \\

& \texttt{\textbf{\textcolor{green}{"lea rdi,[rbx"}}, \textbf{\textcolor{green}{" + rax*1 + 1]\textbackslash n"}}, "jmp addr3\textbackslash n", \textbf{\textcolor{magenta}{"cmp cl,"}}, \textbf{\textcolor{magenta}{"10\textbackslash n"}}, "jnz addr6\textbackslash n",} & \\

& \texttt{\textbf{\textcolor{violet}{"mov word ptr [r"}}, \textbf{\textcolor{violet}{"ax],"}}, \textbf{\textcolor{violet}{"r8"}}, \textbf{\textcolor{violet}{"w\textbackslash n"}}, \textbf{\textcolor{olive}{"add edx,"}}, \textbf{\textcolor{olive}{"2\textbackslash n"}}, "add rsi,1\textbackslash n", \textbf{\textcolor{gray}{"cmp rdi,"}},} & \\

& \texttt{\textbf{\textcolor{gray}{"rsi\textbackslash n"}}, "jz addr4\textbackslash n", \textbf{\textcolor{cyan}{"movzx ecx,byte ptr [r"}}, \textbf{\textcolor{cyan}{"si]\textbackslash n"}}, "movsxd rax,edx\textbackslash n",} & \\

& \texttt{\textbf{\textcolor{pink}{"add rax,"}}, \textbf{\textcolor{pink}{"rbp\textbackslash n"}}, \textbf{\textcolor{teal}{"cmp cl,"}}, \textbf{\textcolor{teal}{"9\textbackslash n"}}, "jnz addr1\textbackslash n", "add rsi,1\textbackslash n",} & \\

& \texttt{\textbf{\textcolor{black}{"mov word ptr [r"}}, \textbf{\textcolor{black}{"ax],"}}, \textbf{\textcolor{black}{"r9"}}, \textbf{\textcolor{black}{"w\textbackslash n"}}, \textbf{\textcolor{orange}{"add edx,"}}, \textbf{\textcolor{orange}{"2\textbackslash n"}}, \textbf{\textcolor{purple}{"cmp rdi,"}}, \textbf{\textcolor{purple}{"rsi\textbackslash n"}},} & \\

& \texttt{"jnz addr3\textbackslash n", "movsxd rdx,edx\textbackslash n", \hl{"add rbp,"}, \hl{"rdx\textbackslash n"}, \textbf{\textcolor{lime}{"mov byte ptr [rbp"}}, \textbf{\textcolor{lime}{"],0\textbackslash n"}},} & \\

& \texttt{"add rsp,8\textbackslash n", "pop rbx\textbackslash n", "pop rbp\textbackslash n", "ret\textbackslash n", \textbf{\textcolor{brown}{"mov byte ptr [rax],"}},} & \\

& \texttt{\textbf{\textcolor{brown}{"cl\textbackslash n"}}, "add edx,1\textbackslash n", "jmp addr2\textbackslash n"} & \\

%%%%%%%%%%%%%%%%%%%%%%%%%%%%%%%%%%%%%%%%%%%%%%%%%%%%%%%%%%%%%%%%%%%%%%%%%%%%%%%%%%%%%%%%%%%%%%%%%%%%%%%%%%%%%%%%%%%%%%%%%%%%%%%%%%%%%%%%%%%%%%%%%%%%%%%%
\hline
\multirow{11}{*}{\textbf{BPE-25K}} & \texttt{"endbr64\textbackslash n", "push rbp\textbackslash n", "mov rbp,rdi\textbackslash n", "mov rdi,rsi\textbackslash n", "push rbx\textbackslash n",} & \multirow{11}{*}{\textbf{49}} \\

& \texttt{"mov rbx,rsi\textbackslash n", "sub rsp,8\textbackslash n", "call addr0\textbackslash n", "test eax,eax\textbackslash n", "jle addr5\textbackslash n",} & \\

& \texttt{"sub eax,1\textbackslash n", \textbf{\textcolor{blue}{"movzx r9d,"}}, \textbf{\textcolor{blue}{"word ptr [addr8]\textbackslash n"}}, "mov rsi,rbx\textbackslash n", "xor edx,edx\textbackslash n",} & \\

& \texttt{\textbf{\textcolor{red}{"movzx r8d,"}}, \textbf{\textcolor{red}{"word ptr [addr7]\textbackslash n"}},\textbf{\textcolor{green}{"lea rdi,[rbx"}}, \textbf{\textcolor{green}{" + rax*1 + 1]\textbackslash n"}}, "jmp addr3\textbackslash n",} & \\

& \texttt{\textbf{\textcolor{magenta}{"cmp cl,10\textbackslash n"}}, "jnz addr6\textbackslash n", \textbf{\textcolor{violet}{"mov word ptr [rax],"}}, \textbf{\textcolor{violet}{"r8w\textbackslash n"}}, \textbf{\textcolor{olive}{"add edx,2\textbackslash n"}},} & \\

& \texttt{"add rsi,1\textbackslash n", \textbf{\textcolor{gray}{"cmp rdi,rsi\textbackslash n"}}, "jz addr4\textbackslash n", \textbf{\textcolor{cyan}{"movzx ecx,byte ptr [rsi]\textbackslash n"}},} & \\

& \texttt{"movsxd rax,edx\textbackslash n", \textbf{\textcolor{pink}{"add rax,rbp\textbackslash n"}}, \textbf{\textcolor{teal}{"cmp cl,9\textbackslash n"}}, "jnz addr1\textbackslash n",} & \\

& \texttt{"add rsi,1\textbackslash n", \textbf{\textcolor{black}{"mov word ptr [rax],"}}, \textbf{\textcolor{black}{"r9w\textbackslash n"}}, \textbf{\textcolor{orange}{"add edx,2\textbackslash n"}}, \textbf{\textcolor{purple}{"cmp rdi,rsi\textbackslash n"}},} & \\

& \texttt{"jnz addr3\textbackslash n", "movsxd rdx,edx\textbackslash n", \hl{"add rbp,rdx\textbackslash n"}, \textbf{\textcolor{lime}{"mov byte ptr [rbp],0\textbackslash n"}},} & \\

& \texttt{"add rsp,8\textbackslash n", "pop rbx\textbackslash n", "pop rbp\textbackslash n", "ret\textbackslash n",} & \\

& \texttt{\textbf{\textcolor{brown}{"mov byte ptr [rax],cl\textbackslash n"}}, "add edx,1\textbackslash n", "jmp addr2\textbackslash n"} & \\
%%%%%%%%%%%%%%%%%%%%%%%%%%%%%%%%%%%%%%%%%%%%%%%%%%%%%%%%%%%%%%%%%%%%%%%%%%%%%%%%%%%%%%%%%%%%%%%%%%%%%%%%%%%%%%%%%%%%%%%%%%%%%%%%%%%%%%%%%%%%%%%%%%%%%%%%
\hline
\multirow{11}{*}{\textbf{BPE-35K}} & \texttt{"endbr64\textbackslash n", "push rbp\textbackslash n", "mov rbp,rdi\textbackslash n", "mov rdi,rsi\textbackslash n", "push rbx\textbackslash n",} & \multirow{11}{*}{\textbf{46}} \\

& \texttt{"mov rbx,rsi\textbackslash n", "sub rsp,8\textbackslash n", "call addr0\textbackslash n", "test eax,eax\textbackslash n",} & \\

& \texttt{"jle addr5\textbackslash n", "sub eax,1\textbackslash n", \textbf{\textcolor{blue}{"movzx r9d,"}}, \textbf{\textcolor{blue}{"word ptr [addr8]\textbackslash n"}}, "mov rsi,rbx\textbackslash n",} & \\

& \texttt{"xor edx,edx\textbackslash n", \textbf{\textcolor{red}{"movzx r8d,word ptr [addr7]\textbackslash n"}}, \textbf{\textcolor{green}{"lea rdi,[rbx + rax*1 + 1]\textbackslash n"}},} & \\

& \texttt{"jmp addr3\textbackslash n", \textbf{\textcolor{magenta}{"cmp cl,10\textbackslash n"}}, "jnz addr6\textbackslash n", \textbf{\textcolor{violet}{"mov word ptr [rax],r8w\textbackslash n"}},} & \\

& \texttt{\textbf{\textcolor{olive}{"add edx,2\textbackslash n"}}, "add rsi,1\textbackslash n", \textbf{\textcolor{gray}{"cmp rdi,rsi\textbackslash n"}}, "jz addr4\textbackslash n",} & \\

& \texttt{\textbf{\textcolor{cyan}{"movzx ecx,byte ptr [rsi]\textbackslash n"}}, "movsxd rax,edx\textbackslash n", \textbf{\textcolor{pink}{"add rax,rbp\textbackslash n"}}, \textbf{\textcolor{teal}{"cmp cl,9\textbackslash n"}},} & \\

& \texttt{"jnz addr1\textbackslash n", "add rsi,1\textbackslash n", \textbf{\textcolor{black}{"mov word ptr [rax],"}}, \textbf{\textcolor{black}{"r9w\textbackslash n"}}, \textbf{\textcolor{orange}{"add edx,2\textbackslash n"}},} & \\

& \texttt{\textbf{\textcolor{purple}{"cmp rdi,rsi\textbackslash n"}}, "jnz addr3\textbackslash n", "movsxd rdx,edx\textbackslash n", \hl{"add rbp,rdx\textbackslash n"},} & \\

& \texttt{\textbf{\textcolor{lime}{"mov byte ptr [rbp],0\textbackslash n"}}, "add rsp,8\textbackslash n", "pop rbx\textbackslash n", "pop rbp\textbackslash n", "ret\textbackslash n",} & \\

& \texttt{\textbf{\textcolor{brown}{"mov byte ptr [rax],cl\textbackslash n"}}, "add edx,1\textbackslash n", "jmp addr2\textbackslash n"} & \\
%%%%%%%%%%%%%%%%%%%%%%%%%%%%%%%%%%%%%%%%%%%%%%%%%%%%%%%%%%%%%%%%%%%%%%%%%%%%%%%%%%%%%%%%%%%%%%%%%%%%%%%%%%%%%%%%%%%%%%%%%%%%%%%%%%%%%%%%%%%%%%%%%%%%%%%%
\hline
\multirow{11}{*}{\textbf{BPE-128K}} & \texttt{"endbr64\textbackslash n", "push rbp\textbackslash n", "mov rbp,rdi\textbackslash n", "mov rdi,rsi\textbackslash n", "push rbx\textbackslash n",} & \multirow{11}{*}{\textbf{44}} \\

& \texttt{"mov rbx,rsi\textbackslash n", "sub rsp,8\textbackslash n", "call addr0\textbackslash n", "test eax,eax\textbackslash n",} & \\

& \texttt{"jle addr5\textbackslash n", "sub eax,1\textbackslash n", \textbf{\textcolor{blue}{"movzx r9d,word ptr [addr8]\textbackslash n"}}, "mov rsi,rbx\textbackslash n",} & \\

& \texttt{"xor edx,edx\textbackslash n", \textbf{\textcolor{red}{"movzx r8d,word ptr [addr7]\textbackslash n"}}, \textbf{\textcolor{green}{"lea rdi,[rbx + rax*1 + 1]\textbackslash n"}},} & \\

& \texttt{"jmp addr3\textbackslash n", \textbf{\textcolor{magenta}{"cmp cl,10\textbackslash n"}}, "jnz addr6\textbackslash n", \textbf{\textcolor{violet}{"mov word ptr [rax],r8w\textbackslash n"}},} & \\

& \texttt{\textbf{\textcolor{olive}{"add edx,2\textbackslash n"}}, "add rsi,1\textbackslash n", \textbf{\textcolor{gray}{"cmp rdi,rsi\textbackslash n"}}, "jz addr4\textbackslash n",} & \\

& \texttt{\textbf{\textcolor{cyan}{"movzx ecx,byte ptr [rsi]\textbackslash n"}}, "movsxd rax,edx\textbackslash n", \textbf{\textcolor{pink}{"add rax,rbp\textbackslash n"}}, \textbf{\textcolor{teal}{"cmp cl,9\textbackslash n"}},} & \\

& \texttt{"jnz addr1\textbackslash n", "add rsi,1\textbackslash n", \textbf{\textcolor{black}{"mov word ptr [rax],r9w\textbackslash n"}}, \textbf{\textcolor{orange}{"add edx,2\textbackslash n"}},} & \\

& \texttt{\textbf{\textcolor{purple}{"cmp rdi,rsi\textbackslash n"}}, "jnz addr3\textbackslash n", "movsxd rdx,edx\textbackslash n", \hl{"add rbp,rdx\textbackslash n"},} & \\

& \texttt{\textbf{\textcolor{lime}{"mov byte ptr [rbp],0\textbackslash n"}}, "add rsp,8\textbackslash n", "pop rbx\textbackslash n", "pop rbp\textbackslash n",} & \\

& \texttt{"ret\textbackslash n", \textbf{\textcolor{brown}{"mov byte ptr [rax],cl\textbackslash n"}}, "add edx,1\textbackslash n", "jmp addr2\textbackslash n"} & \\

\hline
\end{tabular}
\end{table*}
%%%%%%%%%%%%%%%%%%%%%%%%%%%%%%%%%%%%%%%%%%%%%%%%%%%%%%%%%%%%%%%%%%%%%%%%%%%%%%%%%%%%%%%%%%%%%%%%%%%%%%%%%%%%%%%%%%%%%%%%%%%%%%%%%%%%%%%%

\end{document}